\documentclass[10pt,journal,compsoc]{IEEEtran}
\ifCLASSOPTIONcompsoc
  \usepackage[nocompress]{cite}
\else
  \usepackage{cite}
\fi
\usepackage{amsmath,amssymb,amsfonts}
\usepackage{graphicx}
\usepackage{textcomp}
\usepackage{xcolor}

\usepackage{url}
\usepackage{caption}
\usepackage{multirow}
\usepackage{algorithm}
\usepackage{algpseudocode}
\usepackage[caption=false]{subfig}
\usepackage{newfloat}
\usepackage{hyperref}
\hypersetup{
     colorlinks=true,
     linkcolor=black,
     citecolor=black,      
     urlcolor=black,
}

\usepackage{listings}

\definecolor{codegreen}{rgb}{0,0.6,0}
\definecolor{codegray}{rgb}{0.5,0.5,0.5}
\definecolor{codepurple}{rgb}{0.58,0,0.82}
\definecolor{backcolour}{rgb}{0.95,0.95,0.92}

\newcommand\change[1]{\textcolor{black}{#1}}

\lstdefinestyle{mystyle}{
    backgroundcolor=\color{backcolour},   
    commentstyle=\color{codegreen},
    keywordstyle=\color{magenta},
    numberstyle=\tiny\color{codegray},
    stringstyle=\color{codepurple},
    basicstyle=\ttfamily\footnotesize,
    breakatwhitespace=false,         
    breaklines=true,                 
    captionpos=b,                    
    keepspaces=true,                 
    numbers=left,                    
    numbersep=5pt,                  
    showspaces=false,                
    showstringspaces=false,
    showtabs=false,                  
    tabsize=2
}

\newcommand{\batch}{\mathcal{B}}
\newcommand{\weight}{\mathbf{w}}

\ifCLASSINFOpdf
\else
\fi
\begin{document}
%
\title{Scalable K-FAC Training for Deep Neural Networks with Distributed Preconditioning}
%
%
%
%

\author{
\IEEEauthorblockN{Lin Zhang, Shaohuai Shi,~\IEEEmembership{Member,~IEEE}, Wei Wang,~\IEEEmembership{Member,~IEEE}, Bo Li,~\IEEEmembership{Fellow,~IEEE}\\}
\thanks{This paper was produced by the Department of Computer Science and Engineering, The Hong Kong University of Science and Technology. \\
Email: lzhangbv@connect.ust.hk, \{shaohuais, weiwa, bli\}@cse.ust.hk. \\
Corresponding author: Shaohuai Shi (shaohuais@cse.ust.hk). 
}}

\IEEEtitleabstractindextext{%
\begin{abstract}
The second-order optimization methods, notably the D-KFAC (Distributed Kronecker Factored Approximate Curvature) algorithms, have gained traction on accelerating deep neural network (DNN) training on GPU clusters. However, existing D-KFAC algorithms require to compute and communicate a large volume of second-order information, i.e., Kronecker factors (KFs), before preconditioning gradients, resulting in large computation and communication overheads as well as a high memory footprint. In this paper, we propose DP-KFAC, a novel distributed preconditioning scheme that distributes the KF constructing tasks at different DNN layers to different workers. DP-KFAC not only retains the convergence property of the existing D-KFAC algorithms but also enables three benefits: reduced computation overhead in constructing KFs, no communication of KFs, and low memory footprint. Extensive experiments on a 64-GPU cluster show that DP-KFAC reduces the computation overhead by 1.55$\times$-1.65$\times$, the communication cost by 2.79$\times$-3.15$\times$, and the memory footprint by 1.14$\times$-1.47$\times$ \change{in each second-order update} compared to the state-of-the-art D-KFAC methods. 
\end{abstract}

}

\maketitle

\IEEEdisplaynontitleabstractindextext

%
\IEEEpeerreviewmaketitle

\IEEEraisesectionheading{\section{Introduction}\label{sec:intro}}
\IEEEPARstart{I}{n} distributed DNN (deep neural network) training, data parallelism is used as a common practice where the training model is replicated on multiple workers and updated in parallel with local data samples. In the data-parallel settings, a popular training method is distributed synchronous stochastic gradient descent (S-SGD) algorithm~\cite{goyal2017accurate,zhang2017poseidon,jia2018highly,peng2019generic,you2020large}. However, S-SGD only utilizes the first-order gradients to update model parameters, and often requires a large number of iterations to converge~\cite{keskar2016large,hoffer2017train,jia2018highly,shallue2019measuring}, making it slow when training a large model. 

\begin{figure}[!t]
    \centering
    \includegraphics[width=0.8\columnwidth]{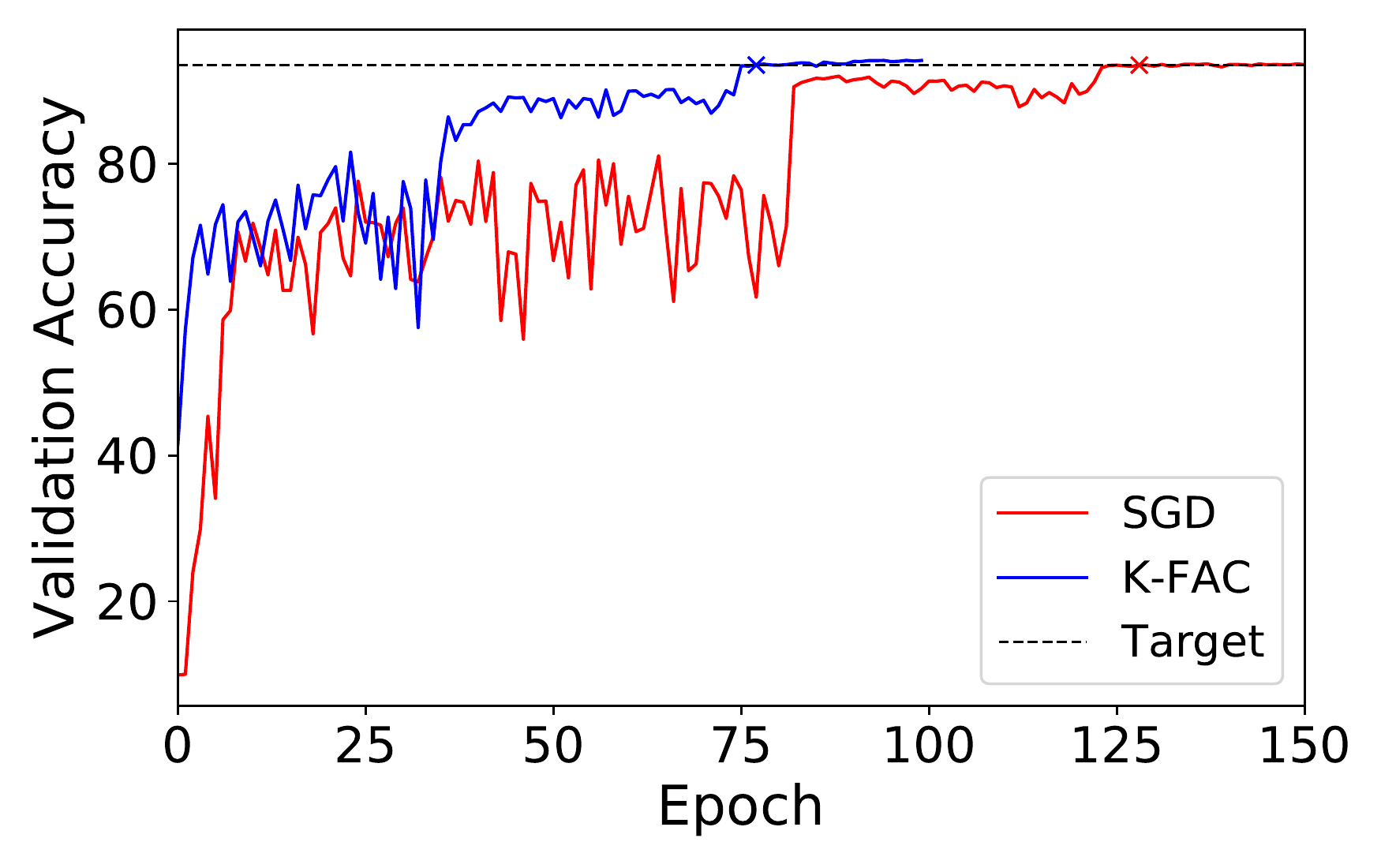}
    \caption{Training the ResNet-110 model on the Cifar-10 dataset with SGD and K-FAC to achieve the target accuracy of 93.5\%. }
    \label{fig:cifar10-resnet110}
\end{figure}

In light of the inefficiency of S-SGD algorithms, second-order optimization methods
have been proposed recently to \textit{precondition} gradients using the Fisher Information Matrix 
(FIM), which are proven effective with evidences in both theory~\cite{amari1998natural,bottou2018optimization,zhang2019fast,martens2020new} and experiments~\cite{osawa2019large,osawa2020scalable,ueno2020rich,pauloski2020convolutional,pauloski2021kaisa,chen2021thor}.
Preconditioning gradient with second-order information often leads to a faster convergence due to the alleviated effect of pathological curvature~\cite{amari2020does}. By using Kronecker Factored Approximate Curvature (K-FAC) algorithms~\cite{martens2015optimizing,grosse2016kronecker,martens2018kronecker,george2018fast}, the FIM can be approximated layer-wise for DNNs in a compute-efficient way. Existing works~\cite{ba2017distributed,osawa2019large,osawa2020scalable,ueno2020rich,pauloski2021kaisa,chen2021thor} have empirically verified that training DNNs with distributed K-FAC (D-KFAC) can converge in a much smaller number of iterations than that of SGD algorithms. Notably, the ResNet-50~\cite{he2016deep} model reaches the target accuracy of 75.9\% on the ImageNet dataset~\cite{deng2009imagenet} in 45 epochs using D-KFAC~\cite{osawa2020scalable,pauloski2021kaisa,chen2021thor}, whereas SGD requires 68 epochs with carefully tuned hyper-parameters according to MLPerf~\cite{mlperf}. Our experimental result (Fig. \ref{fig:cifar10-resnet110}) in training a ResNet-110~\cite{he2016deep} model on the Cifar-10 dataset~\cite{krizhevsky2009learning} also shows that D-KFAC requires 40\% fewer epochs to reach the target 93.5\% validation accuracy~\cite{he2016deep} compared to SGD.

\begin{figure}[!t]
    \centering
    \includegraphics[width=0.8\columnwidth]{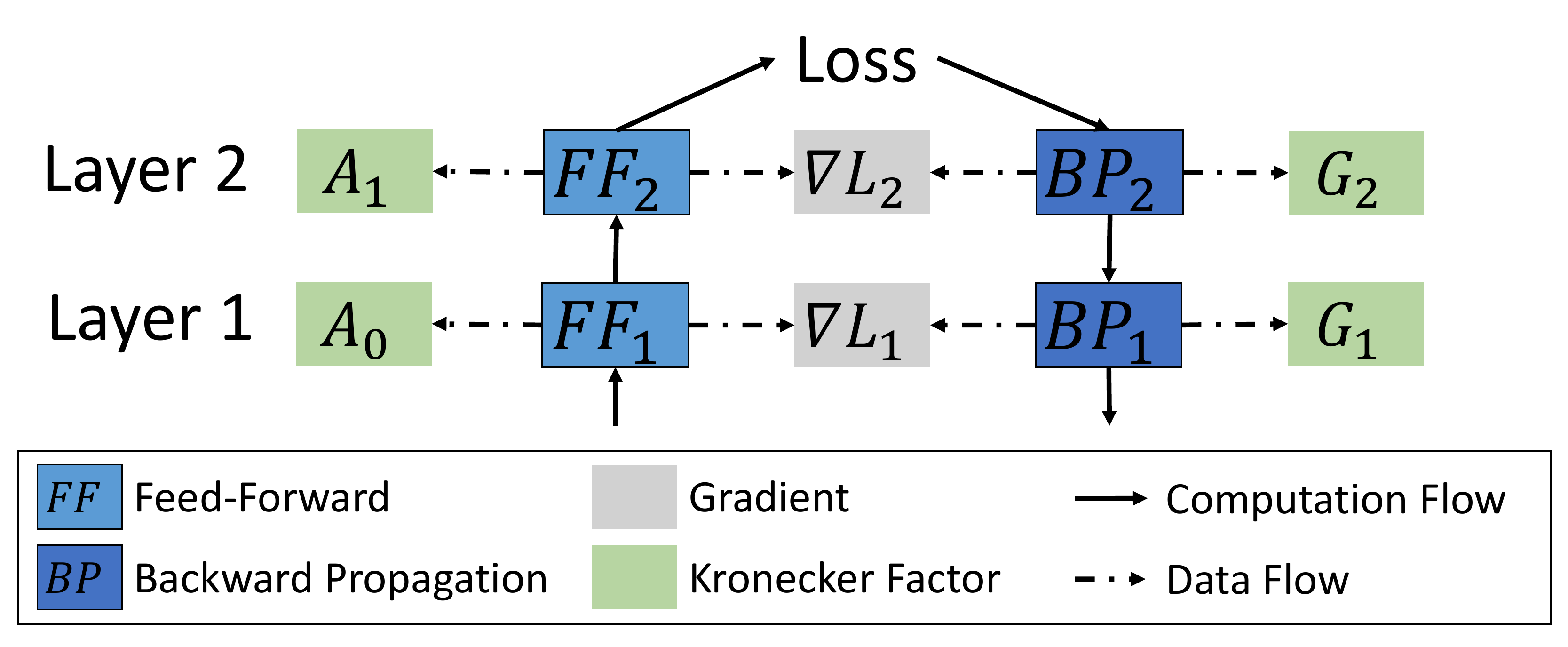}
    \caption{An example of Kronecker factors ($A$ and $G$) computation in a 2-layer DNN.}
    \label{fig:kfac-example}
\end{figure}

Despite the reduced number of iterations, existing D-KFAC algorithms incur expensive overheads in computation, communication, and memory footprint of distributed optimization per iteration, resulting in low scaling efficiency on GPU clusters.
Specifically, in DNN training with SGD, it first samples the data to do the feed-forward (FF) computations to calculate the loss $\mathcal{L}$, and then computes the first-order gradient ($\nabla \mathcal{L}_{i}$ at layer $i$) w.r.t. the model parameters through backward propagation (BP), as shown in Fig.~\ref{fig:kfac-example} (middle). Using K-FAC optimization, it needs to compute the Kronecker factors ($A_{i-1}$ during FF and $G_{i}$ during BP), as shown in Fig.~\ref{fig:kfac-example} (side), to estimate FIMs by $F_i=A_{i-1}\otimes G_{i}$, where $F_i$ is the approximate FIM at layer $i$ of a DNN model and $\otimes$ is the Kronecker product. $F_i$ is then inverted to precondition the gradients, i.e., $F_{i}^{-1} \nabla \mathcal{L}_{i}$. When applying data parallelism with multiple workers, the Kronecker factors (KFs) should be constructed locally and aggregated among workers to generate global KFs which are then inverted to precondition the gradients. This process takes a large portion of the training time in each iteration (more details in \S\ref{sec:inefficiency}). 

In~\cite{osawa2019large,ueno2020rich,osawa2020scalable,pauloski2020convolutional}, the authors propose model parallelism (MPD-KFAC) methods to distributively compute the inverses of KFs, where different GPUs calculate the inverses of different KFs in parallel. From the system's point of view, one can either compute all preconditioned gradients in all workers (which results in minimum communication, called COMM-OPT~\cite{osawa2019large,ueno2020rich,osawa2020scalable,pauloski2021kaisa}) or distributively precondition gradients (which results in minimum temporary results storage, called MEM-OPT~\cite{pauloski2020convolutional,pauloski2021kaisa}) to reduce the iteration time of D-KFAC optimization. Some scheduling algorithms are also proposed to further reduce the iteration time in D-KFAC. For example, SPD-KFAC~\cite{shi2021accelerating} is proposed to pipeline the computations and communications of KFs. In \cite{pauloski2021kaisa}, KAISA is proposed to dynamically determine COMM-OPT and MEM-OPT for particular cluster configurations. In~\cite{chen2021thor}, the KFs update interval can be dynamically determined by their traces. However, all these algorithms require every worker to calculate all layers' KFs which are further aggregated among all workers. Such a requirement introduces significant computation and communication overheads as well as high memory consumption to store KFs. Due to the layer-wise structure of DNNs and layer-wise FIM approximation of K-FAC~\cite{martens2015optimizing}, in this work, we propose a novel \textit{distributed preconditioning} scheme, named DP-KFAC, by preconditioning different layers' gradients using local KFs \change{(partial batch KFs computation)} to accelerate distributed training with little impact on the model convergence. 

In DP-KFAC, we distribute the computing tasks of preconditioning at different layers to different workers, which means each worker only constructs and inverts a part of KFs. Therefore, the per-worker computation workload and memory consumption is significantly reduced, and the communications of KFs among all workers are totally eliminated. We conduct experiments on a 64-GPU cluster connected with 100Gb/s InfiniBand on different applications including CNNs and Transformers. The experimental results show that our DP-KFAC preserves the convergence performance with existing D-KFAC algorithms and it achieves much more efficient training speed than existing state-of-the-arts KAISA~\cite{pauloski2021kaisa} and SPD-KFAC~\cite{shi2021accelerating}. Specifically, on our 64-GPU cluster, DP-KFAC saves 1.55-1.65$\times$ computational costs, 2.79-3.15$\times$ communication costs, 1.14-1.47$\times$ memory consumption, and 1.13-2.27$\times$ iteration time compared to existing state-of-the-art D-KFAC methods KAISA and SPD-KFAC. The main contributions of this paper are summarized as follows:
\begin{itemize}
    \item We systemically analyze the computation and communication overheads of the existing D-KFAC implementations. It is observed that the system bottleneck is caused by the large amount of KFs to be constructed and communicated in distributed optimization.  
    \item We propose a novel distributed preconditioning scheme, named DP-KFAC, to accelerate training by reducing the computation and communication overheads of constructing KFs, and it requires less memory consumption. 
    \item We implement our DP-KFAC system atop existing popular deep learning frameworks PyTorch and Horovod. Users only need to add several lines of code to use DP-KFAC. Codes are available at \url{https://github.com/lzhangbv/kfac_pytorch}. 
    \item We conduct extensive experiments with modern DNNs on a 64-GPU cluster with a 100Gb/s InfiniBand interconnect. The experimental results show that our DP-KFAC is much more efficient in iteration time and consumes less memory storage than state-of-the-art algorithms while maintaining almost the same validation accuracy. 
\end{itemize}

The rest of the paper is organized as follows. We introduce some background and related work in \S\ref{sec:background}. We then analyse the system bottlenecks of the existing D-KFAC implementations in \S\ref{sec:inefficiency}. We present our DP-KFAC design in \S\ref{sec:dp-kfac}, followed by the system implementation in \S\ref{sec:implementation}. We show experimental studies in \S\ref{sec:experiments}. Finally, we conclude the paper in \S\ref{sec:conclusion}. 

\section{Background and Related Work} \label{sec:background}
In DNN training, the target is to minimize a loss function $\mathcal{L}(\weight, D)$, where $\weight$ is the model parameters and $D$ is the training data by iteratively updating $\weight$. It can be optimized with its first-order gradient using SGD or the preconditioned gradient using second-order information. 

\subsection{Distributed SGD}
The mini-batch SGD algorithm updates the model parameters iteratively by using the first-order gradient, i.e.,
\begin{equation}\label{eq:sgd-update}
    \weight^{(t+1)}_i = \weight^{(t)}_i - \alpha^{(t)} \nabla \mathcal{L}_i(\weight^{(t)}, \batch^{(t)}),
\end{equation}
where $\weight^{(t)}_i$ and $\nabla \mathcal{L}_i(\weight^{(t)},\batch^{(t)})$ represent the model parameter and gradient of layer $i$ at iteration $t$ respectively, for $i=1, \cdots, L$ in an $L$-layer DNN. $\alpha^{(t)} > 0$ is the learning rate, and $\batch^{(t)}$ is a mini-batch of data randomly sampled from the training dataset. 

When exploiting data parallelism to train a single model with multiple workers, the distributed synchronized SGD (S-SGD) algorithm updates the model by 
\begin{equation}\label{eq:s-sgd-update}
    \weight^{(t+1)}_i = \weight^{(t)}_i - \alpha^{(t)} \frac{1}{P} \sum_{p=1}^P \nabla \mathcal{L}_i(\weight^{(t)}, \batch^{(t), p}),
\end{equation}
where $\nabla \mathcal{L}_i(\weight^{(t)}, \batch^{(t), p})$ is the local gradient computed on the $p$-th worker with its locally sampled data $\batch^{(t), p}$. Since the local gradients are located among $P$ workers, it is necessary to aggregate the gradients (i.e., summation) before applied to update the model parameters. In the centralized architecture, a parameter server is required to pull local gradients (and push parameters) from (and to) all workers~\cite{li2014scaling}, while in the decentralized architecture, the \textit{all-reduce} collective communication is used to aggregate the gradients among all workers~\cite{goyal2017accurate}. 

\subsection{Distributed K-FAC}
The second-order algorithm with the natural gradient is to use FIMs to precondition the first-order gradient with the following update formula: 
\begin{equation}\label{eq:FIM-update}
    \weight^{(t+1)}_i = \weight^{(t)}_i - \alpha^{(t)} F_i^{-1} \nabla \mathcal{L}_i(\weight^{(t)}, \batch^{(t)}), 
\end{equation}
where $F_i$ is the FIM of layer $i$. The preconditioned gradient, i.e., $F_i^{-1} \nabla \mathcal{L}_i$, often leads to a faster convergence~\cite{martens2015optimizing,bottou2018optimization}, which means it can converge to a solution in fewer number of iterations than SGD. 

To construct the FIM in a more efficient way, K-FAC approximates it as the Kronecker product of two smaller matrices: 
\begin{equation}\label{eq:kronecker-factor-approx}
    F_i \approx \mathbb{E} [a_{i-1} a_{i-1}^T] \otimes \mathbb{E} [g_i g_i^T] \triangleq A_{i-1} \otimes G_{i},
\end{equation}
where $a_{i-1}$ and $g_i$ are the input of layer $i$ (i.e., output at layer $i-1$) and the gradient w.r.t. the pre-activation output of layer $i$, respectively. $A_{i-1}$ and $G_{i}$ are called Kronecker factors (KFs). As shown in Fig.~\ref{fig:kfac-example}, the Kronecker factors $A_{i-1}=\mathbb{E} [a_{i-1} a_{i-1}^T]$ and $G_{i}=\mathbb{E} [g_i g_i^T]$ can be calculated (estimated) during feed-forward and back-propagation with the same mini-batch of data, respectively~\cite{martens2015optimizing}. The update rule of K-FAC in Eq.~(\ref{eq:FIM-update}) can be further represented by 
\begin{equation} \label{eq:kfac-update}
    \weight^{(t+1)}_i = \weight^{(t)}_i - \alpha^{(t)} (A_{i-1}^{-1} \otimes G_{i}^{-1}) \nabla \mathcal{L}_i(\weight^{(t)}, \batch^{(t)}). 
\end{equation}
Compared to SGD, K-FAC needs to construct and invert KFs for preconditioning. 

\begin{figure}[!t]
    \centering
    \includegraphics[width=0.8\columnwidth]{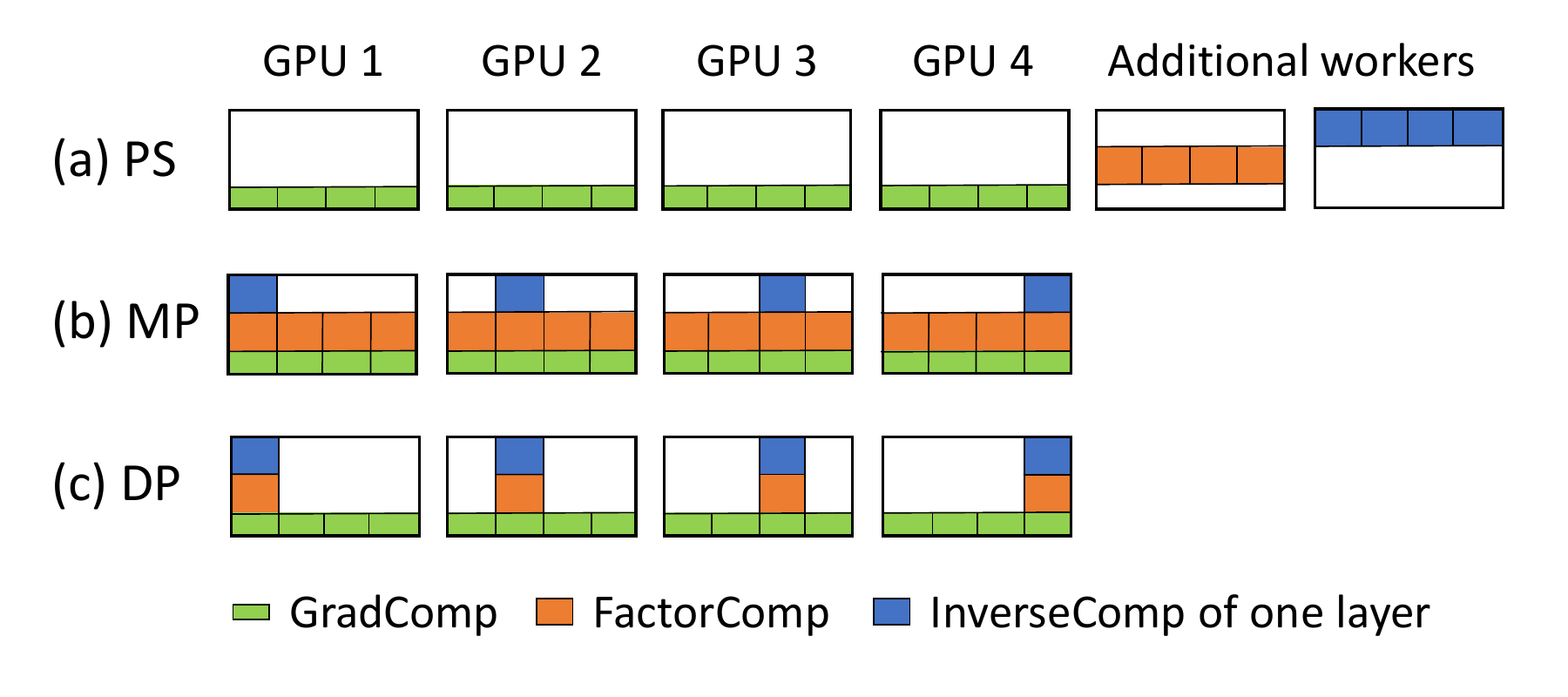}
    \caption{An example of comparing the per-worker computational workloads of gradient computation (GradComp), KFs computation (FactorComp), and inverse computation (InverseComp) with three different D-KFAC implementations (i.e., PS, MP, and our DP), where we specify the number of GPUs as $P=4$ and the number of DNN layers as $L=4$. Any colour chunk denotes a specific computing task of one layer.}
    \label{fig:dkfac-example}
\end{figure}

When exploiting distributed K-FAC (D-KFAC) with data parallelism, the update rule can be represented as follows: 
\begin{align} \label{eq:d-kfac-update}
    \weight^{(t+1)}_i = \weight^{(t)}_i - \alpha^{(t)} (\bar{A}_{i-1}^{-1} \otimes \bar{G}_{i}^{-1}) \nabla \bar{\mathcal{L}}_i,
\end{align}
where $\nabla \bar{\mathcal{L}}_i = \frac{1}{P} \sum_{p=1}^P \nabla \mathcal{L}_i(\weight^{(t)}, \batch^{(t), p})$ is the aggregated gradient at layer $i$ just like S-SGD. $\bar{A}_{i-1}$ and $\bar{G}_{i}$ are the distributed version of KFs used for preconditioning, and their values depend on particular algorithms. 

In the centralized architecture with parameter server (PS)~\cite{li2014scaling}, the distributed K-FAC algorithm (PS-KFAC)~\cite{ba2017distributed} assigns additional workers to calculate and invert KFs locally, as shown in Fig.~\ref{fig:dkfac-example}(a). This means that the computations of gradients and KFs are independent with each other using different data. Once the PS receives inverted KFs from additional workers, they can be directly served as $\bar{A}^{-1}_{i-1}$ and $\bar{G}^{-1}_{i}$ to precondition the aggregated gradient $\nabla \bar{\mathcal{L}}_i$. 

In the decentralized architecture, which is more widely used in practise~\cite{osawa2019large,ueno2020rich,pauloski2021kaisa,chen2021thor}, each worker computes both local gradients and local KFs of all layers; and then the local gradients and KFs are both aggregated via \textit{all-reduce} communications. After \textit{all-reduce}, each worker has the identical aggregated gradient $\nabla \bar{\mathcal{L}}_i$ and global KFs, i.e., $\bar{A}_{i-1}=\frac{1}{P} \sum_{p=1}^P A_{i-1}^p$ and $\bar{G}_{i}=\frac{1}{P} \sum_{p=1}^P G_{i}^p$, where $A_{i-1}^p$ and $G_{i}^p$ are local KFs of layer $i$ constructed on worker $p$. Since the inverse computations of different layers are independent from each other, MPD-KFAC~\cite{osawa2019large,osawa2020scalable,ueno2020rich} adopts the concept of model parallelism (MP) to partition the inverse tasks of different layers to different workers, as shown in Fig.~\ref{fig:dkfac-example}(b). As each worker has only a part of inverse results used for preconditioning, the inverses should be either all-gathered among all workers for preconditioning (i.e., COMM-OPT in~\cite{pauloski2021kaisa}) or used to precondition the gradients locally and then broadcasting the preconditioned gradients to all other workers for model updates (i.e., MEM-OPT in~\cite{pauloski2021kaisa}). Note that most existing D-KFAC algorithms~\cite{osawa2020scalable,ueno2020rich,pauloski2020convolutional,shi2021accelerating} have followed this MP strategy. 

\change{There are also some communication scheduling strategies~\cite{shi2021accelerating,chen2021thor} being proposed to alleviate the communication overheads in D-KFAC through pipelining techniques~\cite{shi2021accelerating} or using dynamic update strategies~\cite{chen2021thor} to improve the scaling efficiency of the distributed system. In this paper, our work is aligned with this direction.}

\section{Analysis of D-KFAC Algorithms}\label{sec:inefficiency}
In this section, we provide an in-depth analysis to the existing D-KFAC algorithms. Specifically, we are interested in the data amount of gradients and KFs, which are directly related to the computation and communication efficiency. 

Let $N_{g}^i$ and $N_{f}^i$ denote the number of elements of the gradient and KFs of layer $i$ ($i=1, \cdots, L$), respectively. If layer $i$ is a simple linear transformation as follows\footnote{Without loss of generality, a convolutional layer can also be formed as a linear transformation~\cite{grosse2016kronecker}.}:
\begin{equation}~\label{eq:linear-layer}
    s_i=W_i a_{i-1} \quad\text{and} \quad a_i = \phi(s_i), 
\end{equation}
where $a_{i-1} \in \mathbb{R}^{d_{i-1}}$ is the input vector, $s_i \in \mathbb{R}^{d_{i}}$ is the pre-activation output vector, $W_i \in \mathbb{R}^{d_{i} \times d_{i-1}}$ is the associated parameter matrix, and $\phi$ is an element-wise activation function. 
Obviously, the gradient $\nabla \mathcal{L}_i$ at this layer has the same number of elements as the parameter $W_i$, that is, $N_{g}^i=d_{i} \times d_{i-1}$. According to Eq.(\ref{eq:kronecker-factor-approx}), KFs $A_{i-1}$ and $G_{i}$ are two square matrices with the dimensions of $d_{i-1}$ and $d_{i}$, respectively. Thus, the total number of elements of KFs in a layer is $N_{f}^{i}=d_{i-1}^2 + d_{i}^2$. Therefore, $\frac{N_{f}^{i}}{N_{g}^{i}} \ge 2$.

\subsection{Computation Overheads}
In the highly optimized D-KFAC design, MPD-KFAC, there are mainly three computing tasks: gradient computation (GradComp), factor computation (FactorComp), and inverse factor computation (InverseComp) as shown in Fig.~\ref{fig:dkfac-example}(b). In the InverseComp stage, each GPU only needs to invert a part of KFs making the overhead of InverseComp be reduced (by $P$ times on average) on a $P$-worker cluster. However, GradComp and FactorComp are still very time-consuming. This is because every worker needs to construct all layers' gradients and KFs, with the total data volume of $N_g=\sum_{i=1}^L N_g^i$ and $N_f=\sum_{i=1}^L N_f^i$, respectively. As $N_f > N_g$, FactorComp typically requires higher workloads than GradComp (including feed-forward and back-propagation). 



\subsection{Communication Overheads}
In terms of the communication tasks, MPD-KFAC consists of gradient communication (GradComm), factor communication (FactorComm), and preconditioned gradient communication (PredComm). In the GradComm and FactorComm stages, the amount of data to be communicated via \textit{all-reduce} operations is $2(P-1) \times N_{g}$ and $2(P-1) \times N_{f}$, for aggregating all local gradients and KFs, respectively. In the PredComm stage, because the preconditioned results of gradients are distributed on different workers, the data amount for broadcasting to all other workers is $(P-1) \times N_{g}$. Therefore, the FactorComm stage has introduced the most expensive communication overheads due to the largest data to be communicated. For example, the ResNet-50~\cite{he2016deep} model has $N_f=153.9$M but $N_g=25.6$M.

\begin{figure}[!ht]
    \centering
    \includegraphics[width=0.7\columnwidth]{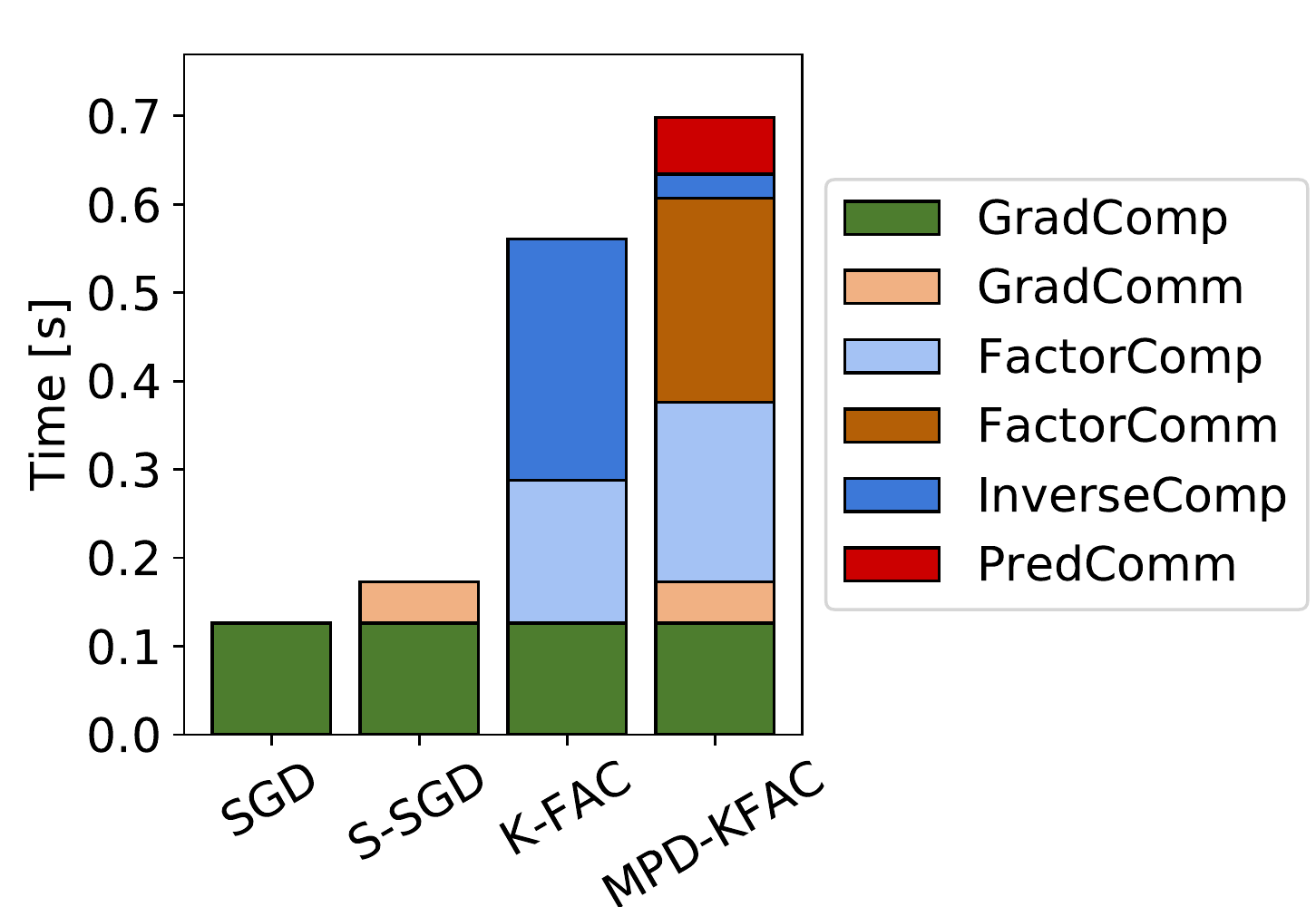}
    \caption{The iteration time breakdowns of training the ResNet-50 model with existing algorithms.}
    \label{fig:naive-breakdowns}
\end{figure}

To illustrate the algorithm efficiency empirically, we provide the iteration time breakdowns of SGD and K-FAC and their distributed versions (S-SGD and MPD-KFAC~\cite{osawa2019large}) by training the ResNet-50 model on a 64-GPU cluster (the details of the system configurations can be found in \S\ref{sec:experiments}). The results are shown in Fig.~\ref{fig:naive-breakdowns}. \change{Since the preconditioning time (PredComp) after the KFs have been inverted is very small compared to parts, we exclude PredComp in Fig.~\ref{fig:naive-breakdowns}.} It is seen that K-FAC runs more than four times slower than SGD at each iteration due to the high costs of constructing and inverting the KFs. MPD-KFAC reduces the inverse computation time vastly using multiple GPUs, but the factor computation and communication overheads become the new bottleneck. 

In summary, existing highly optimized D-KFAC algorithms suffer from the system inefficiency due to the extensive overheads of constructing and communicating KFs of all layers for preconditioning. As a compromise, stale FIM construction and communication~\cite{osawa2019large,chen2021thor} that reduce the frequency of KFs' expensive computations and communications can alleviate the inefficiency. However, skipping FIM updates by utilizing the stale statistics could bring potential negative effects on the convergence performance~\cite{martens2015optimizing,pauloski2020convolutional,pauloski2021kaisa}. In this work, we focus on the per-iteration optimization to tackle the computation and communication challenges in D-KFAC algorithms. 

\section{Distributed Preconditioning}\label{sec:dp-kfac}

As discussed in the previous section, in the FactorComp stage of D-KFAC, each worker (say worker $p$) needs to construct local KFs ($A_{i-1}^p$ during feed-forward) and ($G_i^p$ during back-propagation) with the local sampled data, where $i=1,2,...,L$, for an $L$-layer model. Then the communications are required in the FactorComm stage to aggregate the global KFs $\bar{A}_{i-1}=\frac{1}{P}\sum_{p=1}^P A_{i-1}^p$ and $\bar{G}_{i}=\frac{1}{P}\sum_{p=1}^P G_{i}^p$. This design is based on the assumption that the FIMs should be iteratively estimated according to the current mini-batch of data which is distributed to all participated workers in data parallelism. Thus, all local KFs should be aggregated to generate the global estimates of KFs. On the other hand, in the PS architecture, PS-KFAC~\cite{ba2017distributed} constructs KFs ($A_{i-1}$ and $G_{i}$) with the sampled data different from the data used for calculating the first-order gradients in workers. The results in~\cite{ba2017distributed} show that PS-KFAC converges similarly with the original K-FAC. 

Motivated by PS-KFAC, we propose a \textit{distributed preconditioning} (DP) scheme (i.e., DP-KFAC) for D-KFAC, in the decentralized architecture using data parallelism\change{, which uses a partial view of mini-batch for factor computation.} In our DP-KFAC , each worker only constructs several layers' KFs using its locally sampled data to generate the estimate of empirical FIMs. Then the local KFs are inverted directly on that worker to precondition the corresponding layers' gradients. In other words, KFs of all layers are constructed and inverted distributively instead of using additional workers to calculate and invert KFs in PS-KFAC or aggregating KFs in MPD-KFAC. 

Formally, given an $L$-layer DNN and a $P$-worker cluster, we use $S=\{l_{1}, l_{2}, \cdots, l_{L}\}$ to denote the set of all layers whose gradients need to be preconditioned before updating model parameters. In DP-KFAC, worker $p$ ($p=1,2,...,P$) only constructs KFs in a subset of layers $S_{p} \subset S$ that satisfies
\begin{equation}\label{eq:KFdistribution}
  S_{1}\cup S_{2} \cup ... \cup S_{P}= S,
\end{equation}
and
\begin{equation}
  S_i \cap S_j = \varnothing \text{ for } i \neq j.
\end{equation}

After KFs in $S_{p}$ have been constructed at worker $p$, the worker directly inverts the KFs and then preconditions the first-order gradients on the layers in $S_{p}$. This means that in the FactorComp and InverseComp stages, worker $p$ only computes and inverts the KFs of layers in $S_p$ instead of $S$. As the processes of GradComp and GradComm remain the same as S-SGD in DP-KFAC, each worker has identical aggregated gradients (i.e., $\nabla \bar{\mathcal{L}}_i$ for $i=1,2,...L$) after the \textit{all-reduce} communications. Since worker $p$ only inverts the KFs in $S_p$, the aggregated gradients of layers in $S_p$ are also preconditioned on this worker. Finally, the distributed preconditioned gradients of layers in $S_p$ are broadcast to other workers in the PredComm stage. After that, all workers have the identical preconditioned gradients which are used to update the model parameters. In summary, our DP-KFAC distributes the preconditioning which contains FactorComp and InverseComp stages at different layers to different GPUs (as shown in Fig.~\ref{fig:dkfac-example}(c)) so that the computation and communication overheads can be significantly reduced. 

\begin{figure}[!t]
    \centering
    \includegraphics[width=0.8\columnwidth]{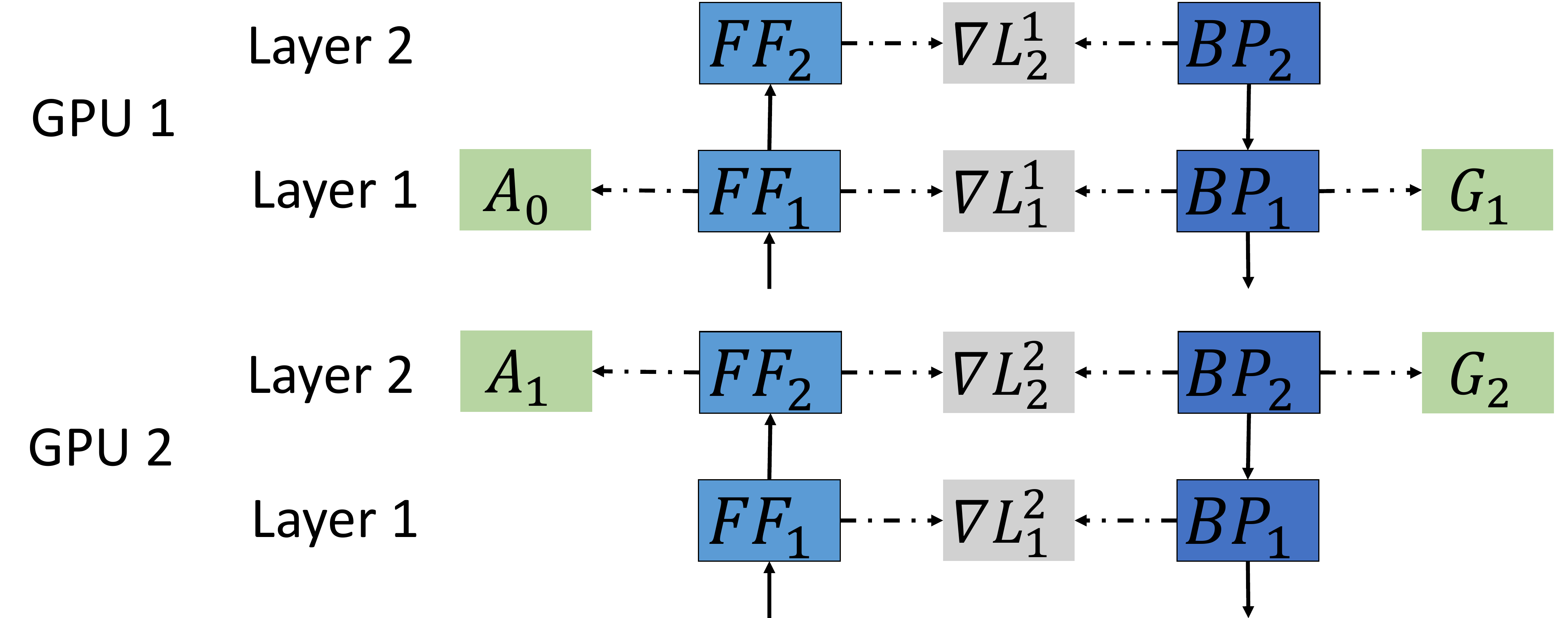}
    \caption{An example of partitioning the tasks of KFs computations to two GPUs in DP-KFAC.}
    \label{fig:partition-factor-comp}
\end{figure}

An example with a 2-layer DNN running on two GPUs with DP-KFAC is shown in Fig.~\ref{fig:partition-factor-comp}. The KFs ($A_{0}$ and $G_{1}$) in layer 1 are calculated on GPU 1, and the KFs ($A_{1}$ and $G_{2}$) in layer 2 are computed on GPU 2. Note that both GPU 1 and GPU 2 compute the local gradients of two layers. This means each GPU only needs to compute two-layer gradients and one-layer KFs. As GPU 1 has the results of $A_{0}$ and $G_{1}$ and GPU 2 has the results of $A_{1}$ and $G_{2}$, they further invert the KFs in parallel. At the same time, the local gradients are communicated via the \textit{all-reduce} collective so that each GPU has the identical aggregated gradients of all layers. After that, GPU 1 can use the results of $A_{0}^{-1}$ and $G_{1}^{-1}$ to precondition the aggregated gradients of layer 1, while the results of $A_{1}^{-1}$ and $G_{2}^{-1}$ are used to precondition the gradients of layer 2 on GPU 2. Therefore, GPU 1 and GPU 2 construct and invert parts of all KFs in parallel.

The pseudo-code of DP-KFAC is shown in Algorithm~\ref{algo:distributed-preconditioning}. At the beginning of training, each worker is assigned with particular layers (line 1) for preconditioning. Lines 2-5 compute and communicate the gradients that are the same with S-SGD. Lines 6-\change{13} precondition the gradients of all layers distributively. In lines \change{14-16}, the distributed preconditioned gradients are broadcast among all workers. Finally, in lines \change{17-19}, each worker updates the model parameters using the identical preconditioned gradients. 
\begin{algorithm}[!ht]
	\caption{DP-KFAC}\label{algo:distributed-preconditioning}
	\begin{algorithmic}[1]
		\small
		\State distribute layers to unique workers in a circular order;
	    \For{each worker $p$}
	        \State compute the gradients $\nabla \mathcal{L}_{1:L}$;
	    \EndFor
	    \State communicate the gradients via \textit{all-reduce}($\nabla \mathcal{L}_{1:L}$);
	    \For{each worker $p$}
	        \For{each layer $i$ assigned to $p$}
	            \State compute the Kronecker factors $A_{i-1}$ and $G_{i}$;
	            \State \change{Running average of $A_{i-1}$ and $G_{i}$;}
	            \State compute the eigen-decompositions or inverses of $A_{i-1}$ and $G_{i}$;
	            \State precondition the gradients; 
	        \EndFor
	    \EndFor
	    \For{each layer $i$ assigned to $p$}
	       \State broadcast the preconditioned gradient from root worker $p$;
	    \EndFor
	    \For{each worker $p$}
	        \State update weights using the preconditioned gradients;
	    \EndFor
	\end{algorithmic}
\end{algorithm}

\subsection{\change{Running Average of KFs}}
In MPD-KFAC, KFs are computed through the global mini-batch size for the purpose of approximating the empirical FIMs over a long-term run~\cite{pauloski2020convolutional}. That is,
\begin{equation} \label{eq:running-average}
    A_{i-1}^{(t)}=\xi A_{i-1}^{(t)} + (1-\xi)A_{i-1}^{(t-1)},
\end{equation}
\begin{equation}
    G_{i}^{(t)}=\xi G_{i}^{(t)} + (1-\xi)G_{i}^{(t-1)},
\end{equation}
where $\xi$ is the running average and typically $\xi\in [0.9, 1)$. Therefore, when using the local mini-batch size in DP-KFAC to compute KFs over a large number of iterations, the expectation of KFs become the same as that with a global mini-batch size. Consequently, different local mini-batch sizes may have little impact on the convergence performance. We further conduct experiments to examine this problem when training with different local mini-batch sizes in \S\ref{sec:experiments}.

\subsection{Complexity analysis}
Our DP-KFAC is mainly composed of three computation stages (GradComp, FactorComp, and InverseComp) and two communication stages (GradComm and PredComm). The GradComp and GradComm stages are the same with S-SGD. Each worker computes $N_g$ elements of gradients, and the amount of data to be communicated is $2(P-1) \times N_g$. For other computation overheads, each worker only needs to construct and invert a fraction of KFs (i.e., $\frac{N_f}{P}$ on average) in the FactorComp and InverseComp stages. In the PredComm stage, the data amount to be broadcast is $(P-1)\times N_g$, which is the same as MPD-KFAC. Therefore, the computation overhead is significantly alleviated by reducing the per-worker workload of FactorComp by $P$ times over MPD-KFAC, and the communication of KFs is totally eliminated. Overall, the efficiency of DP-KFAC with a amount of GPUs becomes comparable to S-SGD. 

In terms of memory consumption, in DP-KFAC, each worker only needs to store $2 \times \frac{N_f}{P}$ elements of KFs of the assigned layers plus their inverse results. Therefore, the memory requirement in DP-KFAC is much less than MPD-KFAC. 

The comparison of complexity between different algorithms is shown in Table~\ref{table:complexity}. 
\begin{table}[!ht]
    \centering
    \caption{Complexity comparison of different algorithms. \change{For MPD-KFAC and DP-KFAC, the complexity is measured on the second-order update.}}
    \label{table:complexity}
    \begin{tabular}{lccc}
    \hline
    Algorithm & S-SGD & MPD-KFAC & DP-KFAC \\ \hline
    GradComp    & $N_g$ & $N_g$ & $N_g$ \\
    FactorComp  & 0 & $N_f$ & $N_f / P$ \\
    InverseComp & 0 & $N_f / P$ & $N_f / P$ \\\hline
    GradComm    & $2(P-1)N_g$ & $2(P-1)N_g$ & $2(P-1)N_g$ \\
    FactorComm  & 0 & $2(P-1)N_f$ & $0$ \\
    PredComm    & 0 & $(P-1)N_g$ & $(P-1)N_g$\\\hline
    Memory      & $N_g$ & $2(N_g+N_f)$ & $2(N_g+N_f/P)$ \\\hline
    \end{tabular}
\end{table}

\section{Implementation}\label{sec:implementation}
We implement our DP-KFAC atop PyTorch~\cite{paszke2019pytorch} and Horovod~\cite{sergeev2018horovod} as PyTorch is a popular deep learning framework and Horovod is widely used for distributed training. \change{To make a minimal change of user code to use our DP-KFAC, the gradient computations and communications are kept unchanged, which are the same with S-SGD implemented in Horovod. \change{Built on the KAISA preconditioner implementation~\cite{pauloski2021kaisa}}, we implement a preconditioner (named ``DP\_KFAC'') to perform distributed preconditioning to the gradient before updating model parameters. Thus, we wrap our DP-KFAC implementation with a PyTorch optimizer class, ``DP\_KFAC''. Unlike KAISA, we support two types of damping (\S\ref{subsec:inversetypes}), matrix-inversion (with ``inv\_type=inverse'') and eigen-decomposition (with ``inv\_type=eigen'') when initializing an instance of ``DP\_KFAC''.}



\subsection{Workload Distribution}
In our implemented DP-KFAC, We apply a round-robin schedule to distribute all layers to different workers in a circular order. For example, given an $L$-layer DNN and a $P$-worker cluster, layers in $S_p=\{l_p, l_{p+P}, l_{p+2P}, \cdots\}$ are assigned to worker $p$. This means that worker $p$ is only required to construct and invert the KFs for preconditioning on a subset of layers $S_p$. In particular, the KFs $A$ and $G$ of layers in $S_p$ are constructed on the worker $p$ during the feed-forward and back-propagation passes by registering the specific hook functions in PyTorch's ``register\_forward\_pre\_hook'' and ``register\_backward\_hook'' APIs, respectively. 

Finally, after the preconditioned gradients have been calculated (\S\ref{subsec:inversetypes}) distributively, we use the \textit{broadcast} collective to collect all results in the PredComm stage before they are used to update the model parameters in all workers. 

\change{A load-balancing workload distribution (e.g., according to the number of floating-point operations per layer) could further improve the training performance. However, its improvement could be much smaller than the time reduced by eliminating the all-reduce operations in aggregating KFs. Thus, we will study this problem in our further version.}

\subsection{Supported Inverse Calculation}\label{subsec:inversetypes}
In the InverseComp stage, the damping strategy~\cite{martens2015optimizing} is often required for preconditioning the gradients with ($F_{i} + \gamma I)^{-1}$, where $I$ is an identity matrix and $\gamma > 0$ is the damping value. The inverse can be computed directly using matrix-inversion by some efficient techniques like Cholesky~\cite{osawa2019large,chen2021thor} or using eigen-decompositions. These two kinds of computations are different in the precondition update formula. 

\textbf{Matrix-inversion. }For the matrix-inversion based damping technique~\cite{martens2015optimizing}, the update formula is given:
\begin{equation}
\label{invere-damped-preconditioning}
    (F_{i}+\gamma I)^{-1} \nabla \bar{\mathcal{L}}_i \change{\approx} (G_i+\frac{\sqrt{\gamma}}{\pi_i}I)^{-1} \change{\nabla} \bar{\mathcal{L}}_i (A_{i-1} + \pi_i \sqrt{\gamma}I)^{-1},
\end{equation}
where $A_{i-1}$ and $G_{i}$ are KFs, $\nabla \bar{\mathcal{L}}_i$ is the aggregated gradient at layer $i$, and $\gamma$ is the damping value. It means that we add KFs by a scaled damping value and then we invert the damped KFs for preconditioning. The scalar constant is given as \change{ $\pi_i=\sqrt{\text{Tr}(A_{i-1})/\text{Dim}(A_{i-1})}/\sqrt{\text{Tr}(G_i)/\text{Dim}(G_i)}$, which minimizes the approximation error of applying damping into KFs, and works better than the choice of $\pi_i=1$ in practice.}

\textbf{Eigen-decomposition. }Regarding the eigen-decomposition based damping technique~\cite{grosse2016kronecker,pauloski2020convolutional}, the update formula should be changed to:
\begin{align} \label{damped-preconditioning}
    (F_{i}+\gamma I)^{-1} \nabla \bar{\mathcal{L}}_i = Q_G \frac{Q^T_G \nabla \bar{\mathcal{L}}_i Q_A}{v_G v_A^T + \gamma} Q^T_A, 
\end{align}
where $A_{i-1}=Q_A \Lambda_A Q_A^T$ and $G_{i}=Q_G \Lambda_G Q_G^T$ are the eigen-decompositions of KFs, $v_A$ and $v_G$ are the diagonals of $\Lambda_A$ and $\Lambda_G$, and $\nabla \bar{\mathcal{L}}_i$ is the aggregated gradient of layer $i$ to be preconditioned. 

In our implementation, we support these two kinds of computations in the preconditioning. By default, we use the eigen-decomposition based damping method to conduct our experiments in the next section because it could provide more stable results~\cite{grosse2016kronecker,pauloski2020convolutional,pauloski2021kaisa}. 

\section{Experimental Studies}\label{sec:experiments}
In this section, we present the experimental studies with several parts. First, we clarify our experimental settings. Second, we compare the convergence performance of DP-KFAC with baselines including S-SGD and MPD-KFAC which is highly optimized in KAISA~\cite{pauloski2021kaisa}, on different applications. Third, we dive into the per-iteration time efficiency and memory consumption of our DP-KFAC compared to state-of-the-arts. Fourth, we discuss the effects of different aspects including the mini-batch size, the number of GPUs, and the frequency of FIM updates. 

\subsection{Experimental Settings}
\textbf{System Configuration.} We conduct our experiments on a 64-GPU cluster. It consists of 16 nodes connected 100Gb/s InfiniBand, and each node has 4 Nvidia RTX2080Ti GPUs (11GB RAM) connected by two Intel(R) Xeon(R) Gold 6230 CPUs, 512GB memory, and PCIe3.0x16. We use some common software including \change{PyTorch-1.10.0}, Horovod-0.21.0, CUDA-10.2, cuDNN-7.6, and NCCL-2.6.4. When running 4-worker experiments, we only use one node of the cluster. 

\textbf{Baselines.} For the first-order algorithms, we use the well-tuned optimizers for particular models, i.e., SGD with momentum for CNNs and Adam~\cite{kingma2014adam} for Transformers. For the D-KFAC variants, we compare our DP-KFAC with KAISA\footnote{The code is adopted from its official implementation at \url{https://github.com/gpauloski/kfac_pytorch}.} with COMM-OPT (KAISA-CO), KAISA with MEM-OPT (KAISA-MO)~\cite{pauloski2021kaisa}, SPD-KFAC~\cite{shi2021accelerating} with pipelining, and THOR~\cite{chen2021thor} with dynamically determining the FIM update frequency. \change{By default, we use KAISA as KAISA-CO if not particularly specified.}

\begin{table}[!ht]
    \centering
     \caption{Hyper-parameter settings. $\alpha$ is the base LR, $P$ is the number of workers, and WU is the number of LR warmup iterations. At decay epochs, LR is decayed by 10 times.}
    \label{table:hypers-lr}
    \begin{tabular}{|c|c|c|c|c|c|}
    \hline
   Model & $\alpha$ & $P$ & WU & Decay Epochs & \# Epochs \\\hline\hline
  ResNet-110 & 0.1 & 4 & 98 & [35, 75, 90] & 100 \\\hline
  VGG-16 & 0.1 & 4 & 98 & [35, 65, 80, 90] & 100 \\\hline
  ResNet-50 & 0.0125 & 64 & 3130 & [25, 35, 40, 45, 50] & 55 \\\hline
  Transformer & 1e-6 & 8 & 4000 & None & 200 \\\hline
  BERT & 5e-6 & 8 & 0 & None & 3 \\\hline
    \end{tabular}
\end{table}
\textbf{DNN Models.} To verify the convergence performance, we choose two modern types of DNNs including CNNs and Transformers similar to the works of KAISA~\cite{pauloski2021kaisa} and THOR~\cite{chen2021thor}. In CNNs, we choose three representative CNN models and datasets, that is ResNet-110 on the Cifar-10~\cite{krizhevsky2009learning} dataset, VGG-16~\cite{simonyan2014very} on the Cifar-100~\cite{krizhevsky2009learning} dataset and ResNet-50 on the ImageNet~\cite{deng2009imagenet} dataset. The Cifar-10 (or Cifar-100) dataset has 10 (or 100) classes with 50,000 training images and 10,000 validation images, and the ImageNet dataset spans 1,000 classes and contains $\sim$1.3M training images and 50,000 validation images. In Transformers, we choose a Transformer~\cite{vaswani2017attention} model on the Multi-30k dataset~\cite{elliott2016multi30k} and a BERT (finetune) \cite{devlin2019bert} model on the SQuAD v1.1 \cite{Rajpurkar2016SQuAD} dataset. The Multi-30k dataset consists of a pair of images and their German and English captions with 29,000 training sentences and 1,014 validation sentences. As this dataset is smaller than the one used in \cite{vaswani2017attention}, we instead use a shallow Transformer model with only two hidden layers as \cite{shallue2019measuring} to achieve better German-to-English translation results. The SQuAD represents a standard question answering dataset with more than 100K crowdsourced question-answer pairs. Since training BERT from scratch is extremely time-consuming, we choose to finetune a pre-trained BERT model on SQuAD using the Transformers library \cite{Wolf2020Transformers}.

\textbf{Hyper-parameters. }
For all algorithms (S-SGD and D-KFAC), we adopt the learning rate (LR) scheduling strategy from~\cite{goyal2017accurate}. Specifically, we use the linear scaling rule that multiplies a base LR ($\alpha$) by the number of workers ($P$) when applied data parallelism to train DNN models. In early stages of training (say warmup), we start from the base LR of $\alpha$ and enlarge it by a constant amount at each iteration such that it reaches $P \times \alpha$. After warmup, the LR decays by
$1/10$ multiple times at given epochs for better convergence. All hyper-parameters that relate to LR are listed in Table~\ref{table:hypers-lr}. For other hyper-parameters, we use moving average parameters for gradients (say momentum) and KFs~\cite{martens2015optimizing} of 0.9 and 0.95, respectively. We use S-SGD+ to denote the SGD (or Adam) algorithm that runs more epochs to reach the target accuracy, i.e., we train ResNet-110 and VGG-16 for 165 epochs (LR decays at epochs 82 and 123), and ResNet-50 for 90 epochs (LR decays at epochs 30, 60, and 80), Transformer for 400 epochs, and BERT for 6 epochs. For D-KFAC related algorithms, we use the damping $\gamma$ with 0.03 for different models, except 0.002 for ResNet-50. 

\subsection{Convergence Performance}
\begin{figure*}[!t]
    \centering
    \subfloat[VGG-16 on Cifar-100]{
        \includegraphics[width=0.48\columnwidth]{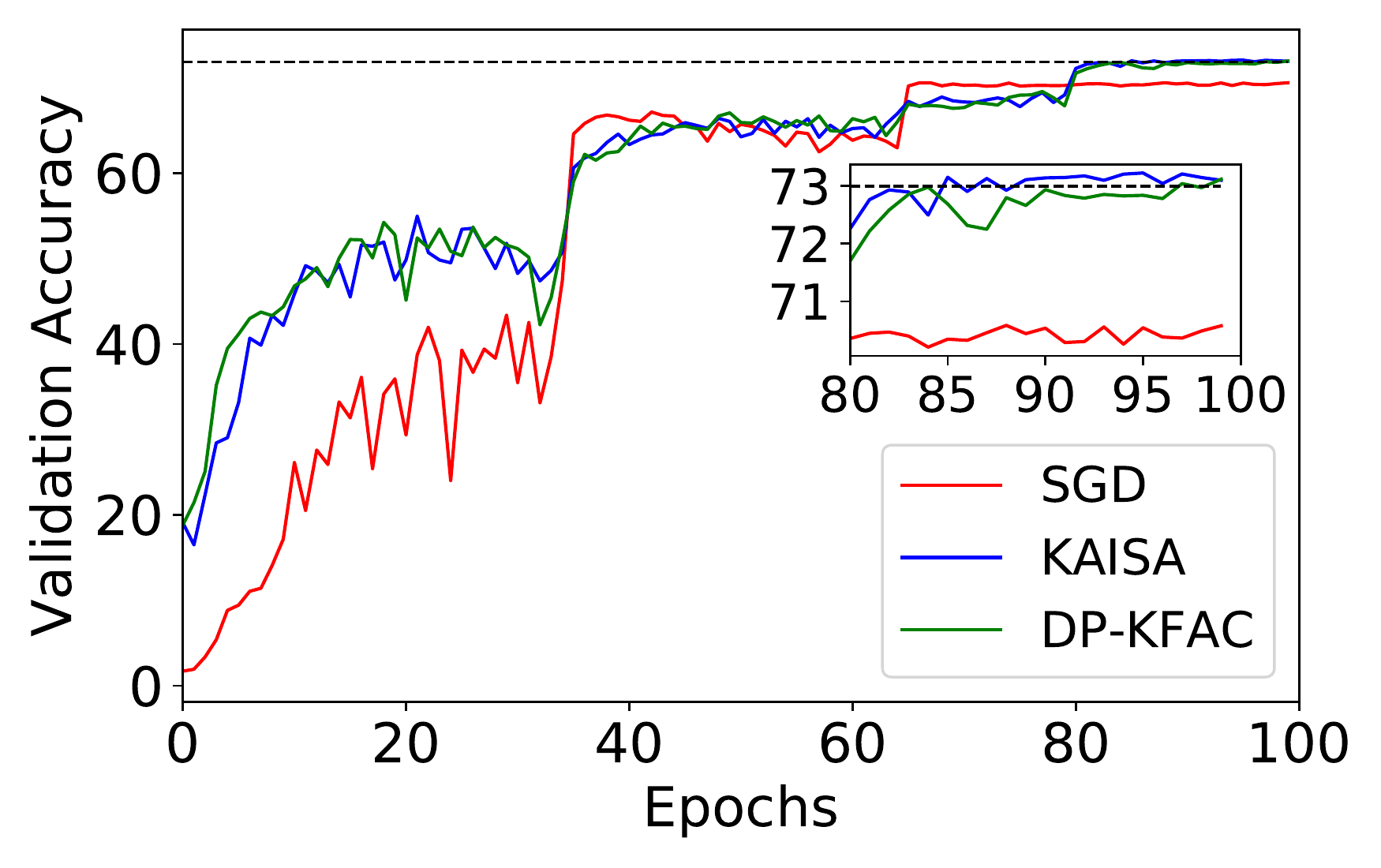}
    }
    \subfloat[ResNet-50 on ImageNet]{
        \includegraphics[width=0.48\columnwidth]{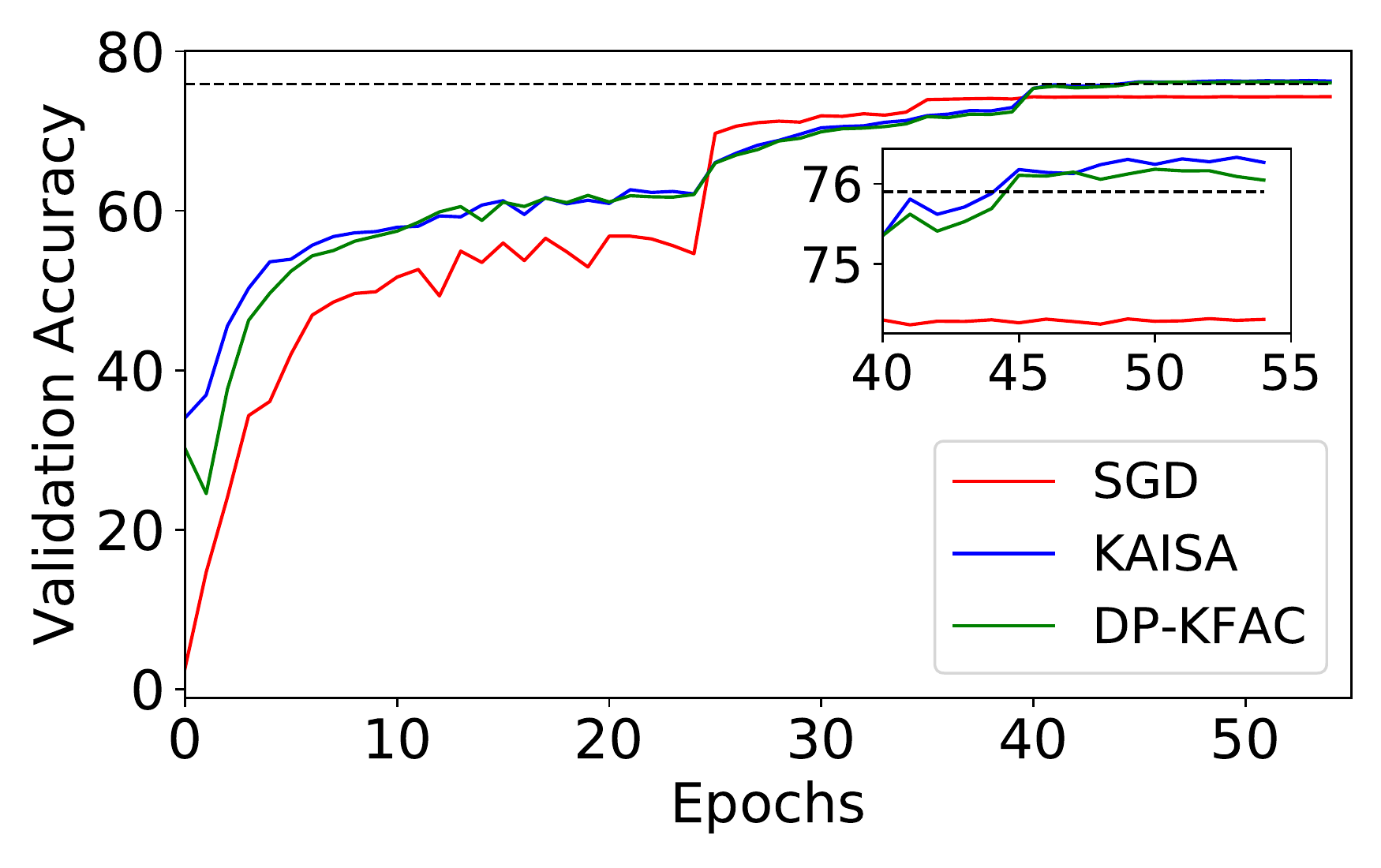}
    } ~
    \subfloat[Transformer on Multi-30k]{
        \includegraphics[width=0.48\columnwidth]{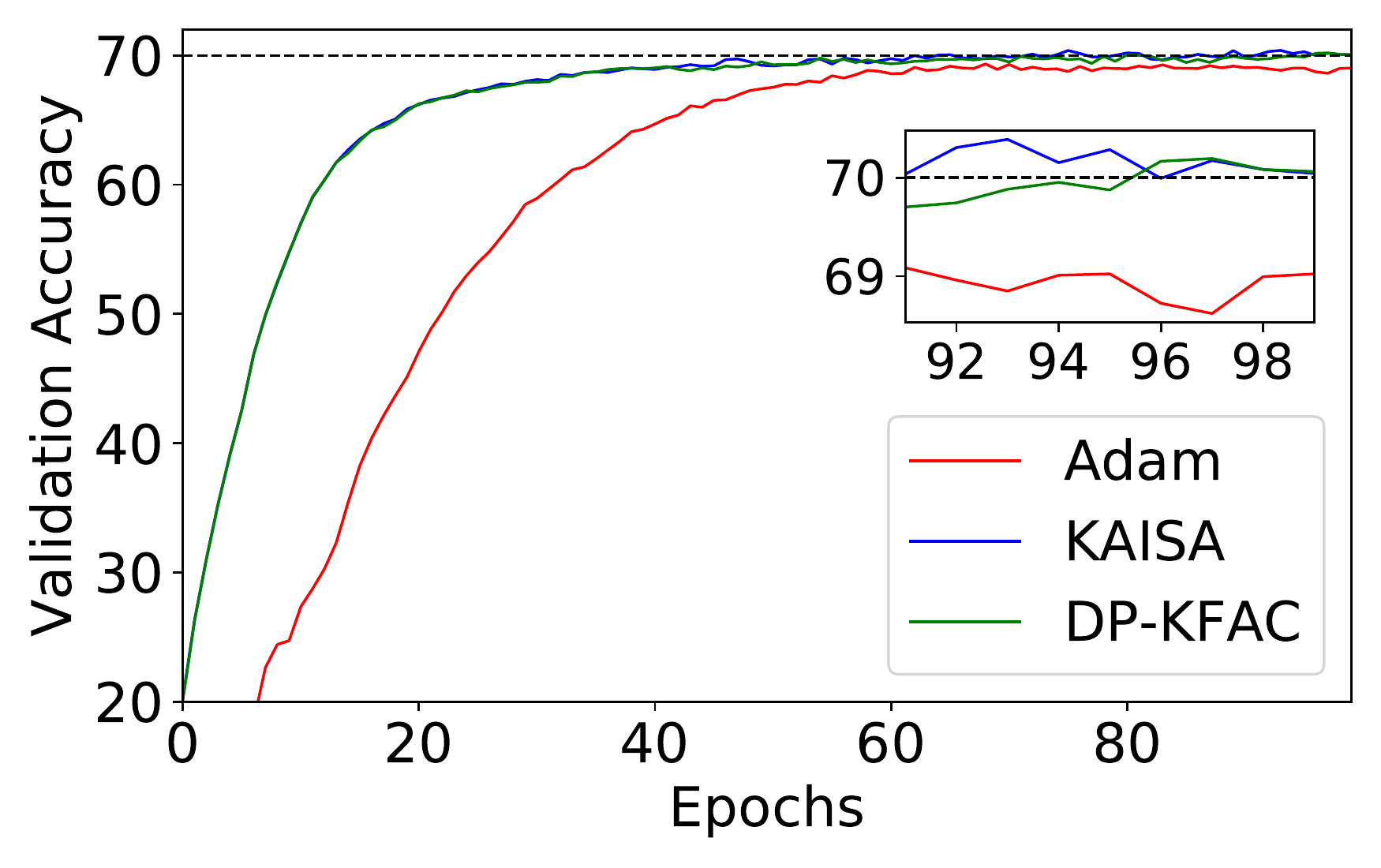}
    } ~
    \subfloat[BERT on SQuAD]{
        \includegraphics[width=0.48\columnwidth]{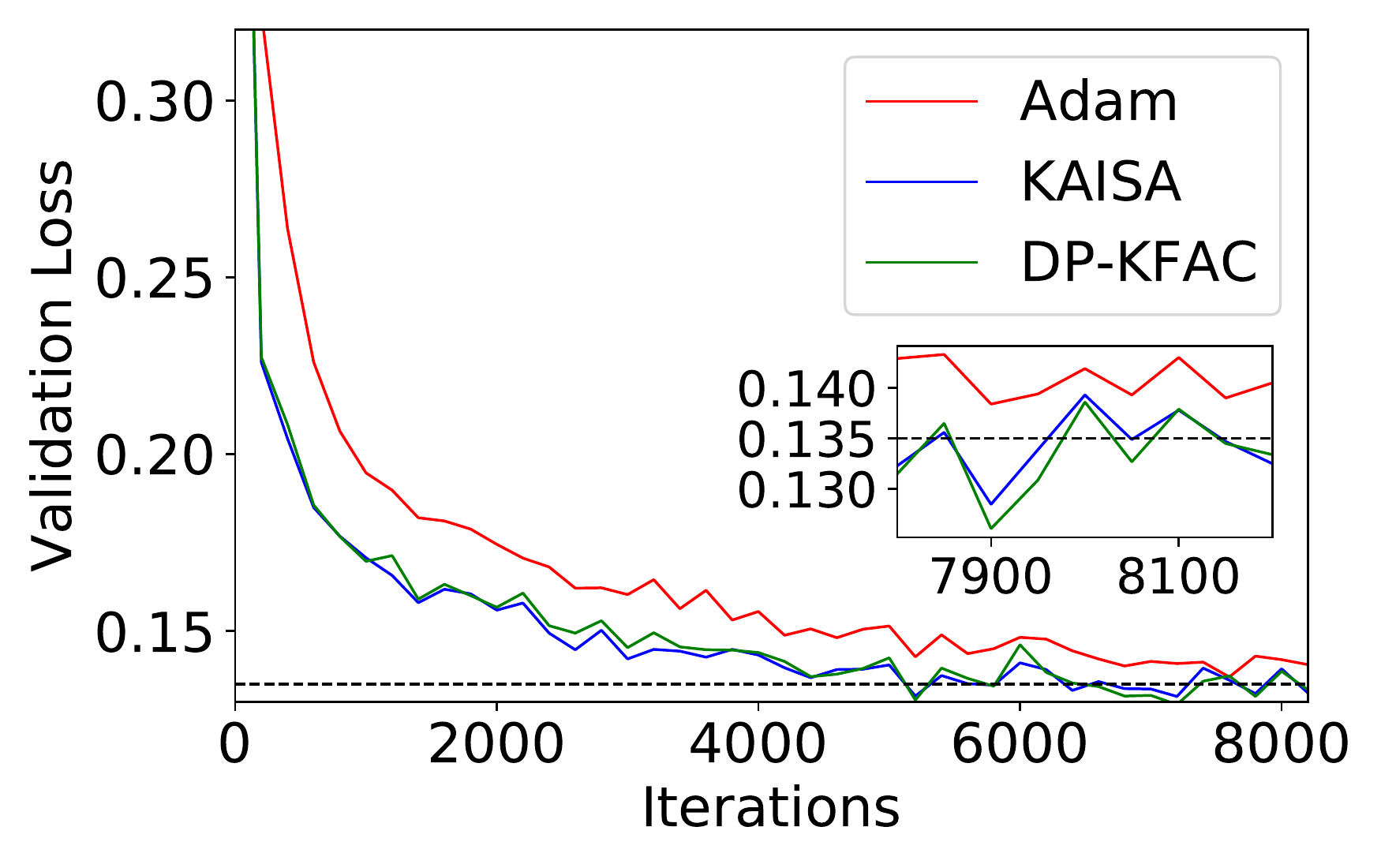}
    }
    \caption{Convergence performance comparison between DP-KFAC and SGD/KAISA on different DNN models and datasets. The black dashed line represents the target accuracy. }
    \label{fig:convergence}
\end{figure*}
We compare the convergence performance of our DP-KFAC with KAISA\footnote{Since KAISA and THOR are with the same computation and communication schemes, and their main difference is the update interval of the FIMs, they should have similar convergence performance as shown in~\cite{chen2021thor}. So we only run KAISA to compare the convergence.} and S-SGD (with momentum) or Adam. For each application, the preconditioners are applied in convolution layers and linear layers in both KAISA and DP-KFAC.

\textbf{CNNs. }
The convergence curves (top-1 validation accuracy vs. epochs) on CNNs are shown in Fig.~\ref{fig:convergence}(a) and (b). The results indicate that DP-KFAC converges almost the same with KAISA, and they both outperform S-SGD at the same number of epochs. In VGG-16 running 100 epochs, both DP-KFAC and KAISA achieve the target accuracy of 73\% at around 90 epochs, while S-SGD only converges to 70.58\%. In ResNet-50, DP-KFAC and KAISA achieve the target accuracy of 75.9\% at epoch 45, while S-SGD only converges to 74.31\%. As expected, since the second-order methods can converge faster than the first-order methods, S-SGD needs to run more iterations to achieve the target accuracy than the K-FAC methods. The final converged accuracy is shown in Table~\ref{table:accuracy}, where S-SGD, KAISA, and DP-KFAC run with the same number of epochs as shown in Table~\ref{table:hypers-lr} while S-SGD+ runs extra epochs (165 epochs on ResNet-110 and VGG-16, and 90 epochs on ResNet-50). The results show that our DP-KFAC achieves very close model accuracy with KAISA, and both of them can achieve the target accuracy in a much fewer number of iterations than SGD.

\textbf{Transformers. } We report the convergence curves (validation accuracy vs. epochs, and validation loss vs. iterations) on Transformer and BERT models in Fig.~\ref{fig:convergence}(c) and Fig.~\ref{fig:convergence}(d), respectively. The results validate that DP-KFAC converges as fast as KAISA on Transformer-based models for different NLP tasks, while Adam converges slower than K-FAC algorithms, and achieves poorer performance with the same training iterations. For final converged results, we use the BLEU score and F1 score to measure the quality of model predictions in translation and question answering tasks, respectively. As shown in Table ~\ref{table:accuracy}, DP-KFAC and KAISA achieve very close BLEU scores (38.66 and 38.47) on the Multi-30k de-en dataset, both of which are higher than the results of Adam (36.69) and S-SGD+ (37.15). On the SQuAD dataset, DP-KFAC and KAISA have almost the same F1 scores (87.86 and 87.73), which is better than Adam (86.05) and S-SGD+ (87.16). From above results, it has shown that K-FAC algorithms can achieve better convergence performance than the first-order counterpart in both CV and NLP domains, and our DP-KFAC algorithm can preserve the model accuracy compared to the KAISA without distributed preconditioning. 

\begin{table}[!t]
    \centering
     \caption{Best validation accuracy of different optimization algorithms under the configured epochs as shown in Table~\ref{table:hypers-lr}. S-SGD+ indicates that the models are trained with S-SGD (or Adam) using more epochs to improve model accuracy. The top-1 accuracy (\%) is used to measure CNNs, while BLEU and F1 metrics are used to measure Transformer and BERT, respectively. We measure 3 times for each experiment and report their mean and std. }
    \label{table:accuracy}
    \centering
    \addtolength{\tabcolsep}{-0.4pt}
    \begin{tabular}{|c|c|c|c|c|}
    \hline
   Model & S-SGD & S-SGD+ & KAISA & DP-KFAC \\\hline\hline
  ResNet-110 & 93.25$\pm$0.08 & 93.66$\pm$0.09 & 93.89$\pm$0.12 & \textbf{93.99$\pm$0.10} \\\hline
  VGG-16 & 70.44$\pm$0.13 & \textbf{73.23$\pm$0.10} & 73.21$\pm$0.09 & 73.12$\pm$0.08 \\\hline
  ResNet-50 & 74.31$\pm$0.10 & \textbf{76.42$\pm$0.03} & 76.10$\pm$0.06 & 76.34$\pm$0.12 \\\hline
  Transformer & 36.69$\pm$0.05 & 37.15$\pm$0.05 & 38.47$\pm$0.26 & \textbf{38.66$\pm$0.18} \\\hline
  BERT & 86.05$\pm$0.03 & 87.16$\pm$0.06 & 87.73$\pm$0.24 & \textbf{87.86$\pm$0.07} \\\hline
    \end{tabular}
\end{table}

\subsection{Training Efficiency}
\begin{table*}[!ht]
    \centering
     \caption{Average \change{second-order} iteration time (in seconds) and occupied GPU memory (in GB) measured by 100 iterations.}
    \label{table:timeandmem}
    \centering
    \begin{tabular}{|c|c|c|c|c|c|c|c|c|c|c|}
    \hline
  \multirow{2}{*}{Model} & \multirow{2}{*}{Dataset} & \multirow{2}{*}{\# GPUs} & \multicolumn{2}{c|}{KAISA-CO} & \multicolumn{2}{c|}{KAISA-MO} & \multicolumn{2}{c|}{SPD-KFAC} & \multicolumn{2}{c|}{DP-KFAC} \\\cline{4-11}
        & &                     &  Time & Mem & Time & Mem & Time & Mem & Time & Mem \\\hline\hline
  ResNet-110 & Cifar-10 &  4    & 0.865$\pm$0.071 & 3.3  & 0.840$\pm$0.027 & 3.2 & 0.743$\pm$0.008 &  3.4 & \textbf{0.514$\pm$0.007} & \textbf{3.2} \\\hline
  VGG-16 & Cifar-100    & 4     & 1.394$\pm$0.007 & 4.3  & 1.305$\pm$0.008 & 4.0 &   1.207$\pm$0.006 & 4.9 & \textbf{1.062$\pm$0.005} & \textbf{3.5} \\\hline
  ResNet-50  & \multirow{3}{*}{ImageNet} & \multirow{3}{*}{64} & 1.375$\pm$0.010 & 8.8 & 1.204$\pm$0.002 & 8.0 & 1.125$\pm$0.005 & 9.8 & \textbf{0.744$\pm$0.002} & \textbf{7.0} \\\cline{1-1}\cline{4-11}
  DenseNet-201  &&& 2.030$\pm$0.011 & 9.3 &  1.624$\pm$0.011 & 8.1 & 1.609$\pm$0.005 & 10.7 & \textbf{0.894$\pm$0.008} & \textbf{6.6} \\\cline{1-1}\cline{4-11}
  Inception-v4  &&& 1.703$\pm$0.005 & 7.5 & 1.402$\pm$0.011 & 6.5 & 1.305$\pm$0.008 & 8.8 & \textbf{0.803$\pm$0.003} & \textbf{5.1} \\\hline
  Transformer & Multi-30k &  8  & 1.390$\pm$0.016 & 7.6  & 1.274$\pm$0.017 & 7.3 & 1.348$\pm$0.028 &  8.3 & \textbf{1.049$\pm$0.033} & \textbf{6.8} \\\hline
  BERT & SQuAD &  8  & 1.812$\pm$0.009 & 8.1  & 1.644$\pm$0.005 & 7.1 & 1.703$\pm$0.009 &  8.9 & \textbf{1.211$\pm$0.007} & \textbf{6.1} \\\hline
    \end{tabular}
\end{table*}

As we have shown the convergence performance of our DP-KFAC, we would like to demonstrate its per-iteration training efficiency and memory consumption compared to existing highly optimized solutions including KAISA-CO, KAISA-MO~\cite{pauloski2021kaisa}, and SPD-KFAC~\cite{shi2021accelerating}. Specifically, KAISA-CO and SPD-KFAC broadcast the inverses of KFs and then compute the preconditioned gradients in all workers, while KAISA-MO and DP-KFAC precondition the gradients distributively and then broadcast the preconditioned gradients among all workers. 
Except the models listed in Table~\ref{table:hypers-lr}, we select two more popular CNNs, DenseNet-201~\cite{huang2017densely} and Inception-v4~\cite{szegedy2017inception}, on the ImageNet dataset and the per-GPU mini-batch size is set as 16 to fully utilize GPU memory. 

The per-iteration time and GPU memory consumption are shown in Table~\ref{table:timeandmem}. The results show that our DP-KFAC outperforms all the other state-of-the-art algorithms in per-iteration time and GPU memory consumption in all tested models. Particularly, on the relatively small models, ResNet-110 and VGG-16, our DP-KFAC runs 14\%-68\% faster than KAISA (including both KAISA-CO and KAISA-MO) and SPD-KFAC. On the large-scale models with ImageNet, our DP-KFAC significantly reduces the iteration time by 1.5$\times$-2.3$\times$ compared to KAISA and SPD-KFAC on the 64-GPU cluster. Regarding the Transformers, DP-KFAC runs 21\%-50\% faster than KAISA and SPD-KFAC. In terms of occupied GPU memory, DP-KFAC consumes 26\%, 15\%, and 42\% less memory than KAISA-CO, KAISA-MO, and SPD-KFAC. Note SPD-KFAC consumes the most memory with tensor fusion to reduce the iteration time. In addition, we notice KAISA-MO outperforms KAISA-CO in both time and memory costs, this is because we consider the per-iteration optimization, in which the iterative communication overhead of KAISA-MO (PredComm) is smaller than KAISA-CO (InverseComm). \change{Because KAISA-CO is motivated by using stale FIM, which reduces the frequency of InverseComm, we will study the effects of the update frequency of KFs in the discussions.}

\subsection{Time Breakdowns}
\begin{figure}[!ht]
    \centering
    \includegraphics[width=0.9\columnwidth]{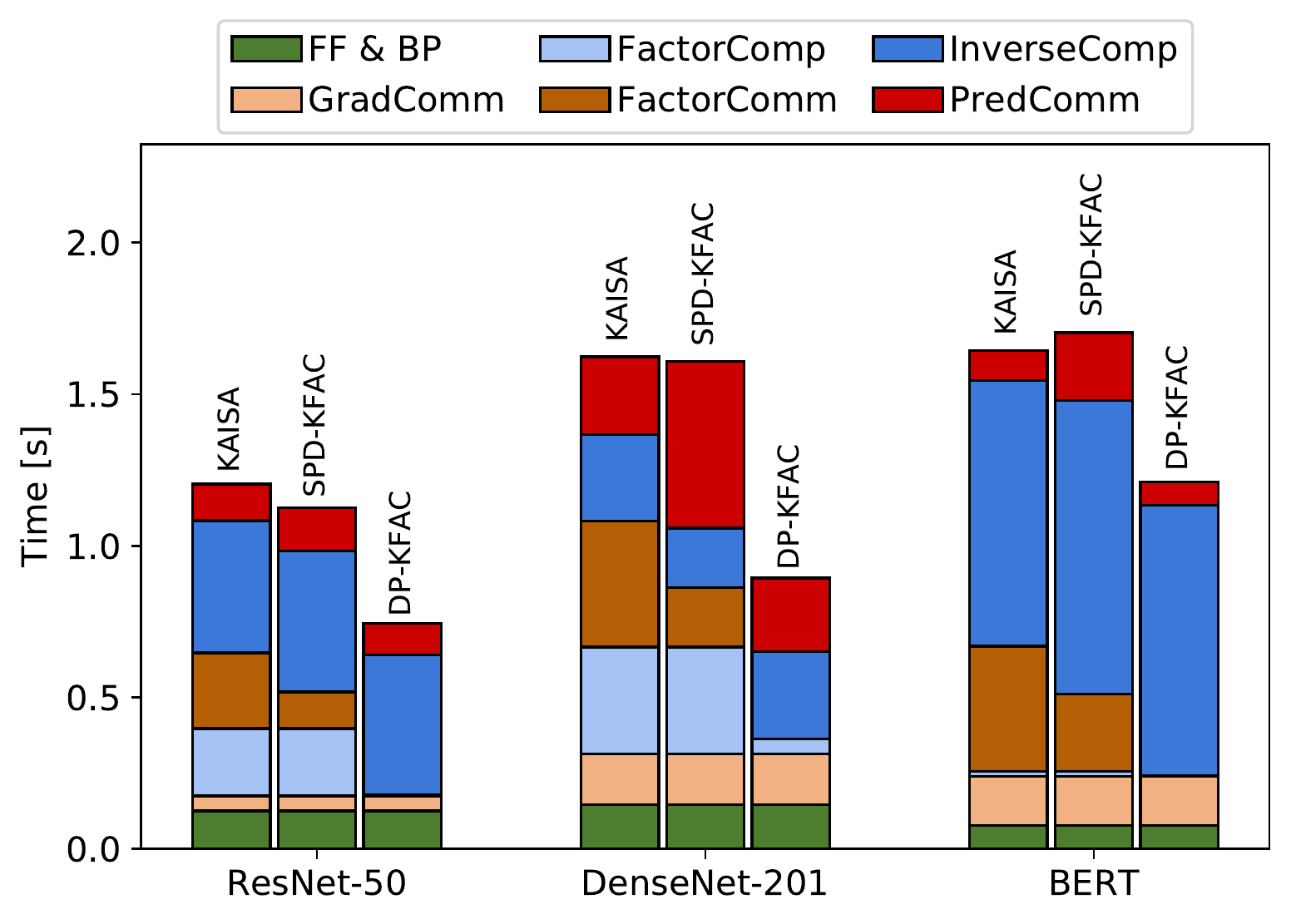}
    \caption{Time breakdown comparison between different D-KFAC algorithms with eigen-decomposition based damping. }
    \label{fig:time-breakdown-kfac-eigen}
\end{figure}

\change{\textbf{Time Performance.}} To understand the time consumption in different D-KFAC algorithms, We dive into the time breakdowns of one training iteration between KAISA-MO, SPD-KFAC, and DP-KFAC, on three representative models: ResNet-50, DenseNet-201, and BERT. The results are given in Fig. \ref{fig:time-breakdown-kfac-eigen}. 

First, the feed-forward and back-propagation computations (FF\&BP) and gradient communications (GradComm) are the same on different algorithms, because the S-SGD part is implemented based on Horovod and it is independent from the K-FAC preconditioner. Second, in terms of KF computations (FactorComp), the FactorComp cost in DP-KFAC is significantly reduced compared to KAISA and SPD-KFAC. Note that the cost of FactorComp on the BERT is small on all algorithms since the batch size of BERT for estimating KFs is only 4. Third, for KF communications (FactorComm), though SPD-KFAC can reduce some FactorComm overheads by overlapping the computation and communication tasks, our DP-KFAC can totally eliminate FactorComm. Hence, DP-KFAC has zero FactorComm cost and small FactorComp cost, which contribute to performance improvement of DP-KFAC. However, the InverseComp cost could be the bottleneck even when the eigen-decomposition computations are distributively performed in DP-KFAC. One alternative is to use the matrix-inversion damping strategy instead, as inverting damped KFs is more time efficient than eigen-decomposition computations. 

\begin{figure}[!ht]
    \centering
    \includegraphics[width=0.9\columnwidth]{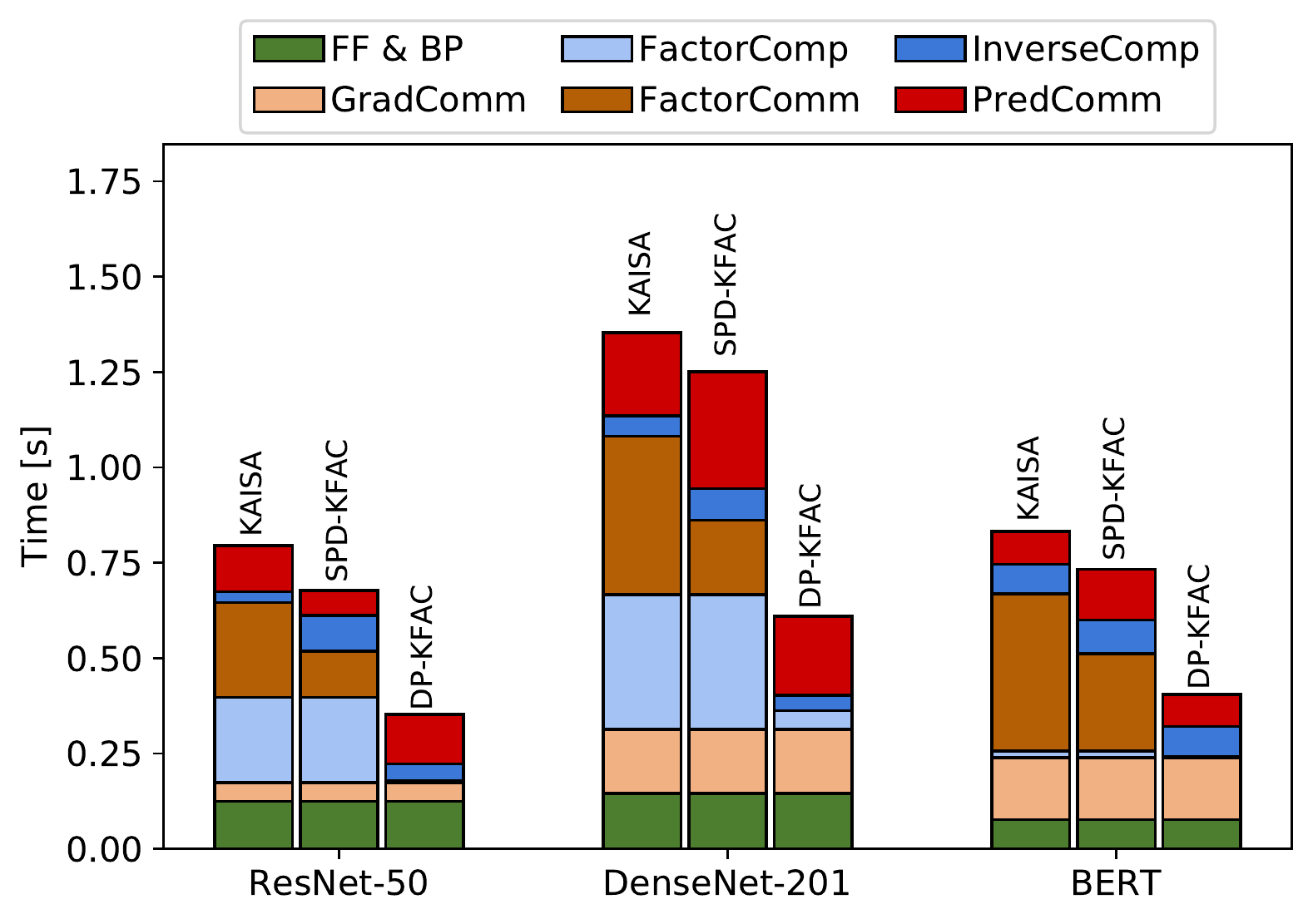}
    \caption{Time breakdown comparison between different D-KFAC algorithms with matrix-inversion based damping. }
    \label{fig:time-breakdown-kfac-inverse}
\end{figure}

Therefore, we provide time breakdowns of matrix-inversion damping based D-KFAC algorithms in Fig. \ref{fig:time-breakdown-kfac-inverse} with the same settings. It shows that the matrix-inversion based damping can in average reduce the iteration time of DP-KFAC by 2.2x compared to those using eigen-decomposition based damping. Besides, DP-KFAC outperforms KAISA and SPD-KFAC as it can also reduce the FactorComp cost and remove the FactorComm overhead. Furthermore, using matrix-inversion in InverseComp makes the iteration time of DP-KFAC be much closer to that of S-SGD than using eigen-decomposition. \change{Specifically, the running time of the second-order iteration of DP-KFAC is around 1.7$\times$-2$\times$ slower than the first-order iteration time of S-SGD.}

\change{\textbf{Convergence Performance.} Besides, we provide the validation accuracy of KAISA (inv) and DP-KFAC (inv) with matrix-inversion based damping on different datasets following the same configurations as shown in Table~\ref{table:hypers-lr}. The results are given in Table~\ref{table:accuracy-mat-inv}. It shows that DP-KFAC (inv) can maintain the validation accuracy over different datasets compared to KAISA (inv) without distributed preconditioning. As DP-KFAC (inv) can achieve the target accuracy as well as DP-KFAC (eigen), it is suggested to use DP-KFAC (inv) to further accelerate the training process with lower computation cost.} \change{The matrix-inversion based K-FAC algorithms achieved worse validation accuracy compared to using eigen-decomposition~\cite{pauloski2020convolutional,pauloski2021kaisa}, but in this work, we find that the scalar $\pi_i$ (see Eq.~\ref{invere-damped-preconditioning}) is helpful to stabilize the training process. Apart from that, we tune the running average value $\xi$ (see Eq.~\ref{eq:running-average}), e.g., set $\xi=0.05$ on Cifar-100, for better performance.}

\begin{table}[!t]
    \centering
     \caption{\change{Best validation accuracy of KAISA and DP-KFAC with eigen-decomposition or matrix-inversion based damping. }}
    \label{table:accuracy-mat-inv}
    \centering
    \addtolength{\tabcolsep}{-0.4pt}
    \begin{tabular}{|c|c|c|c|c|}
    \hline
   Algorithm & ResNet-110 & VGG-16 & Transformer & BERT \\\hline\hline
   KAISA (eigen) & 93.89 & 73.21 & 38.47 & 87.73 \\\hline
   DP-KFAC (eigen) & 93.99 & 73.12 & 38.66 & 87.86 \\\hline
   KAISA (inv) & 93.83 & 73.26 & 38.17 & 87.75 \\\hline
   DP-KFAC (inv) & \textbf{94.35} & \textbf{73.27} & \textbf{38.90} & \textbf{88.00} \\\hline
    \end{tabular}
\end{table}

\subsection{Discussions} \label{subsec:discussion}
In this subsection, we would like to discuss the learning properties of our DP-KFAC algorithm with different aspects, including the per-GPU batch size, the number of GPUs, the damping value, and the frequency of FIM updates. We finally compare the end-to-end training time of DP-KFAC to SGD and existing D-KFAC algorithms by training the ResNet-50 model with stale FIMs.

\textbf{Effects of local mini-batch sizes.}
In MPD-KFAC, KFs are computed through the global mini-batch size for the purpose of approximating the empirical FIMs over a long-term run. Intuitively, when using the local mini-batch size in DP-KFAC to compute KFs over a large number of iterations, the expectation of KFs become the same as that with a global mini-batch size. Consequently, different local mini-batch sizes may have little impact on the convergence performance. We further conduct experiments (100 epochs) to examine this problem when training with different local mini-batch sizes \change{compared with SGD and KAISA as shown in Table~\ref{table:batchsize}}. \change{The results show that: 1) DP-KFAC achieves comparable convergence performance with KAISA under different mini-batch sizes; 2) Both DP-KFAC and KAISA significantly outperform SGD with a fixed number of iterations, especially on VGG-16; 3) With an increased mini-batch size, the convergence performance tends to decrease in all algorithms, which could be related to the generalization gap in large-batch training~\cite{keskar2016large,lin2020extrapolation}.} \change{In distributed training, data may be partitioned across multiple nodes, which makes each node have only a part of data for calculating gradients~\cite{nguyen2022globally}. It may also have some impacts on the convergence on SGD~\cite{nguyen2022globally} and KFAC algorithms. We will leave this as our future study.}


\begin{table}[!ht]
    \centering
     \caption{Validation accuracy (\%) with different local mini-batch sizes in \change{SGD, KAISA, and} DP-KFAC running on 4 GPUs. We measure 3 times for each experiment and report their mean and std. \change{Bold texts are the best validation accuracy among the compared algorithms.}}
    \label{table:batchsize}
    \centering
    \addtolength{\tabcolsep}{-2.5pt}
    \begin{tabular}{|c|c|c|c|c|c|}
    \hline
   \multirow{2}{*}{Model} & \multirow{2}{*}{Algorithm} & \multicolumn{4}{c|}{Local mini-batch size} \\\cline{3-6}
    & & 32 & 64 & 128 & 256 \\\hline\hline
    \multirow{3}{*}{ResNet-110} & SGD & 93.78$\pm$.14 & 93.43$\pm$.12 & 93.04$\pm$.48 & 91.31$\pm$.37 \\\cline{2-6}
        & KAISA & 93.84$\pm$.11 &93.95$\pm$.14 & 94.0$\pm$.14& \textbf{93.79$\pm$.16} \\\cline{2-6}
        & DP-KFAC & \textbf{94.03$\pm$.17} & \textbf{94.04$\pm$.17} & \textbf{94.02$\pm$.08} & 93.79$\pm$.20 \\\hline
    \multirow{3}{*}{VGG-16} & SGD & 72.14$\pm$.12 & 71.67$\pm$.45 &70.32$\pm$.49 & 69.54$\pm$.51 \\\cline{2-6}
        & KAISA & 72.94$\pm$.39 & \textbf{73.5$\pm$.27} & \textbf{73.21$\pm$.09} &72.67$\pm$.21 \\\cline{2-6}
        & DP-KFAC & \textbf{73.02$\pm$.30} & 73.34$\pm$.20 & 73.12$\pm$.08 & \textbf{72.82$\pm$.48} \\\hline 
    \end{tabular}
\end{table}

\begin{figure}[!t]
    \centering
    \subfloat[ResNet-110 on Cifar-10]{
        \includegraphics[width=0.48\columnwidth]{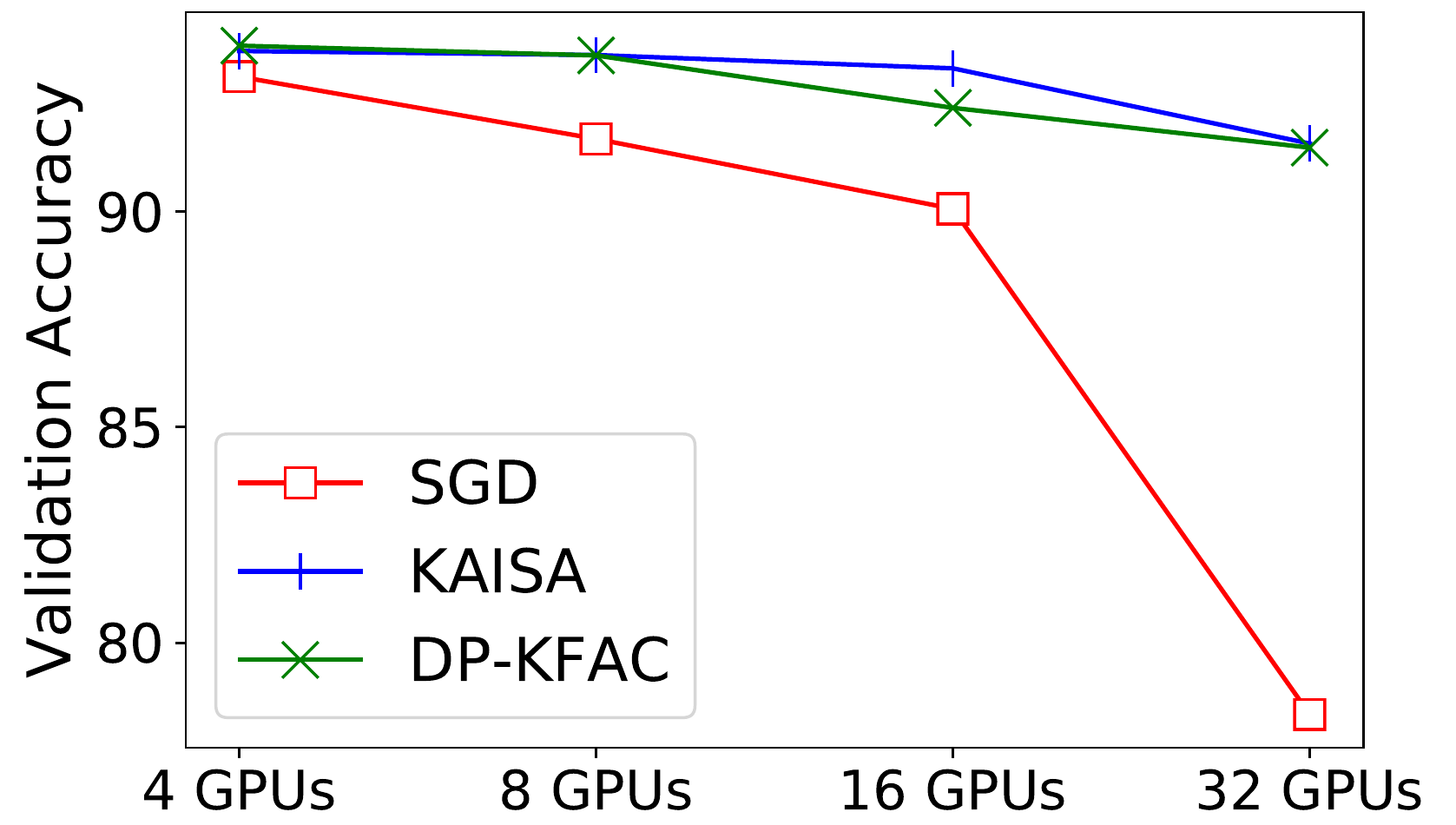}
    }
    \subfloat[VGG-16 on Cifar-100]{
        \includegraphics[width=0.48\columnwidth]{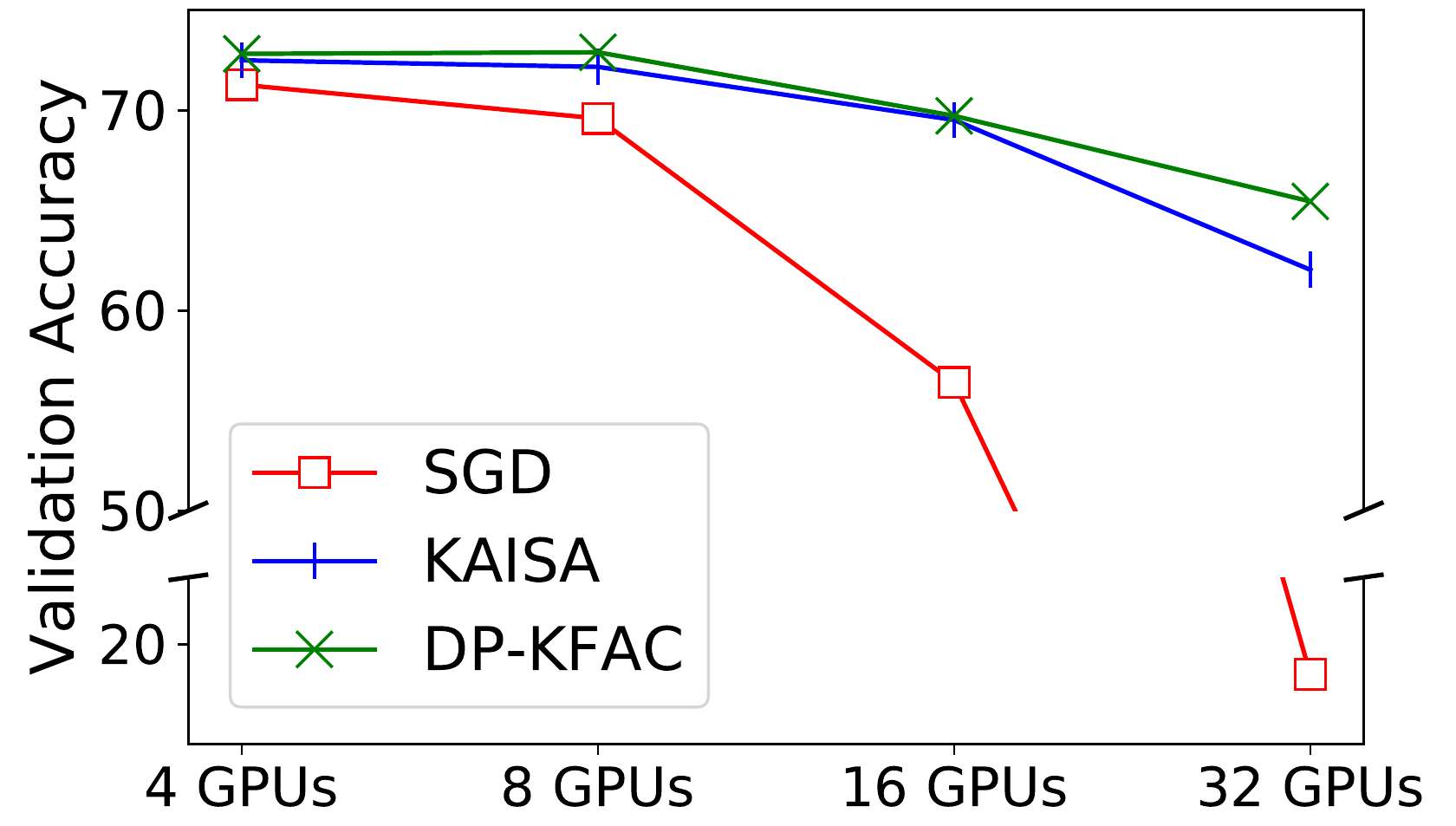}
    }
    \caption{Validation accuracy comparison between DP-KFAC and SGD/KAISA on ResNet-110 and VGG-16 with different number of GPUs. }
    \label{fig:convergence-nworkers}
\end{figure}

\textbf{Effects of the number of GPUs in convergence.}
The key communication and computation workloads reduction in our DP-KFAC is to estimate the KFs with local mini-batch data and construct the FIMs distributively without factor communications, while MPD-KFAC algorithms estimate the KFs with global mini-batch data. Since the global mini-batch size is $P$ times larger than local mini-batch size, we hereby report the training performance when the number of workers $P$ changes. We run SGD, KAISA, and DP-KFAC on ResNet-110 and VGG-16 models, with different number of GPUs. We use a damping $\gamma$ of 0.001 to avoid possible divergence, and keep all other hyper-parameters fixed. The validation accuracy results are shown in Figure~\ref{fig:convergence-nworkers}. The results indicate that our DP-KFAC achieves comparable performance to KAISA on two models with different number of GPUs. As expected, they both outperform SGD in all tested cases. As the number of GPUs increases, the performance of all algorithms with the same epochs drops in different degrees. This is because the number of iterations is also reduced with large data parallelism~\cite{goyal2017accurate}. The empirical results show that D-KFAC could scale better to larger batch size than SGD. 

\begin{figure}[!ht]
    \centering
    \subfloat[ResNet-50]{
        \includegraphics[width=0.48\columnwidth]{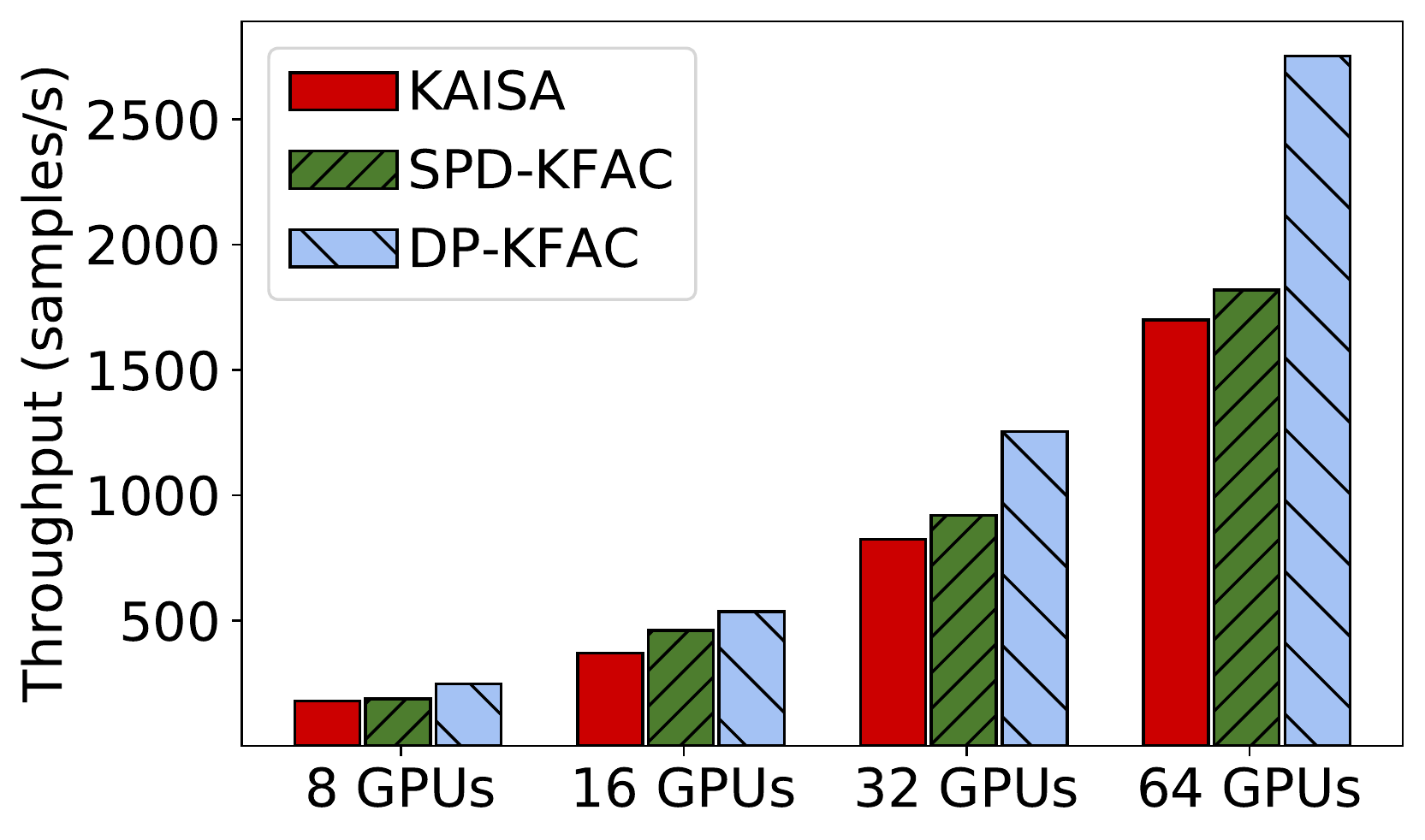}
    }
    \subfloat[DenseNet-201]{
        \includegraphics[width=0.48\columnwidth]{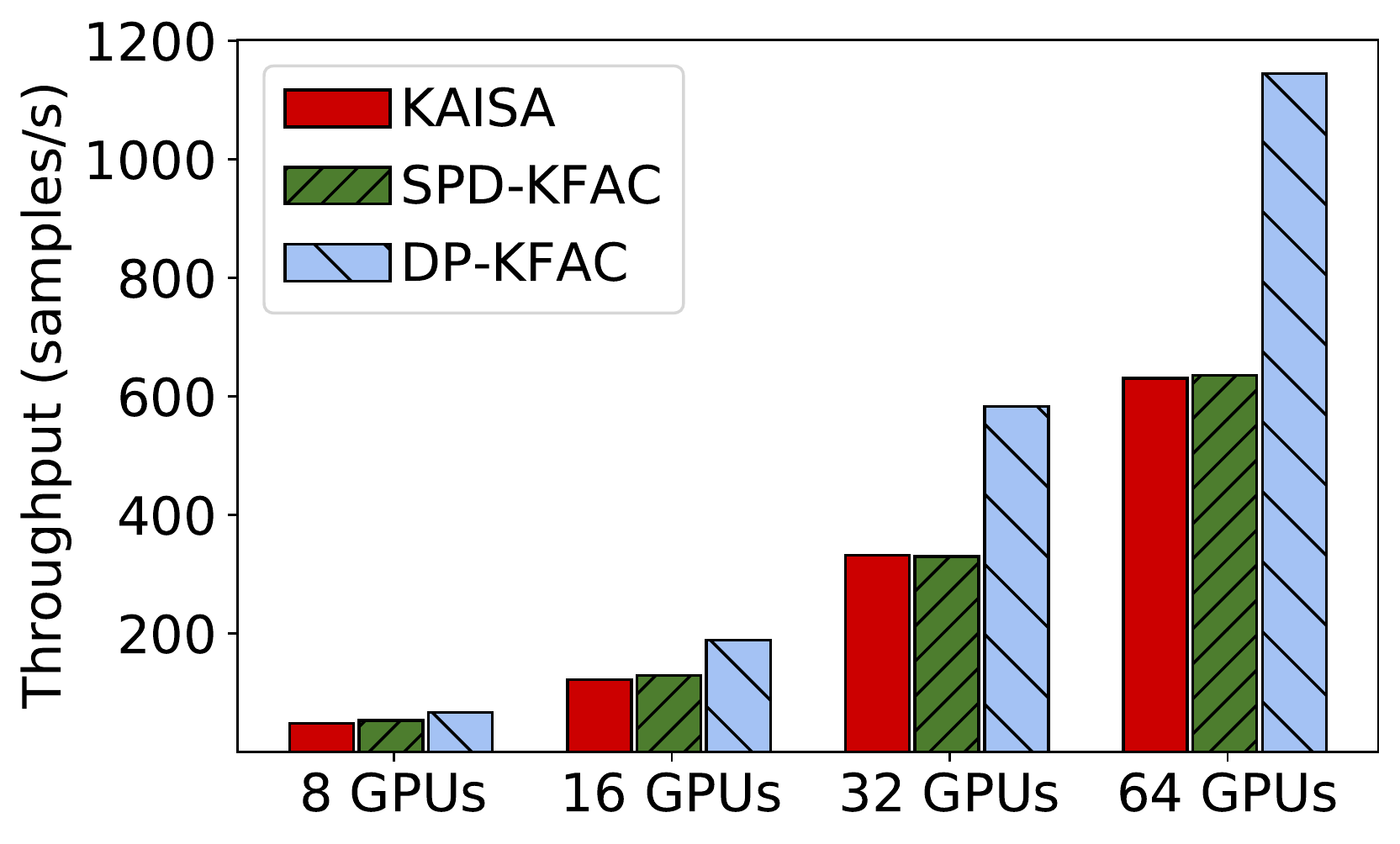}
    }
    \caption{System throughput comparison between DP-KFAC and KAISA/SPD-KFAC on the ImageNet dataset. }
    \label{fig:system_throughput}
\end{figure}

\textbf{Effects of the number of GPUs in throughput.}
To validate the scalability of DP-KFAC when the number of GPUs changes, we also compare the system throughput on ResNet-50 and DenseNet-201 models. The results are shown in Figure \ref{fig:system_throughput}, indicating that our DP-KFAC can achieve the best system throughput in all tested cases, and it scales much better than others when the number of GPUs increases. It is seen that with more GPUs, our DP-KFAC has higher benefits than others since the time cost of data aggregation of KFs is proportional to the number of workers in both SPD-KFAC and KAISA, but DP-KFAC has no need to communicate KFs.

\textbf{Effects of damping and the update frequency of KFs. }
\begin{figure}[!ht]
    \centering
    \subfloat[KAISA, 4 GPUs]{
        \includegraphics[width=0.48\columnwidth]{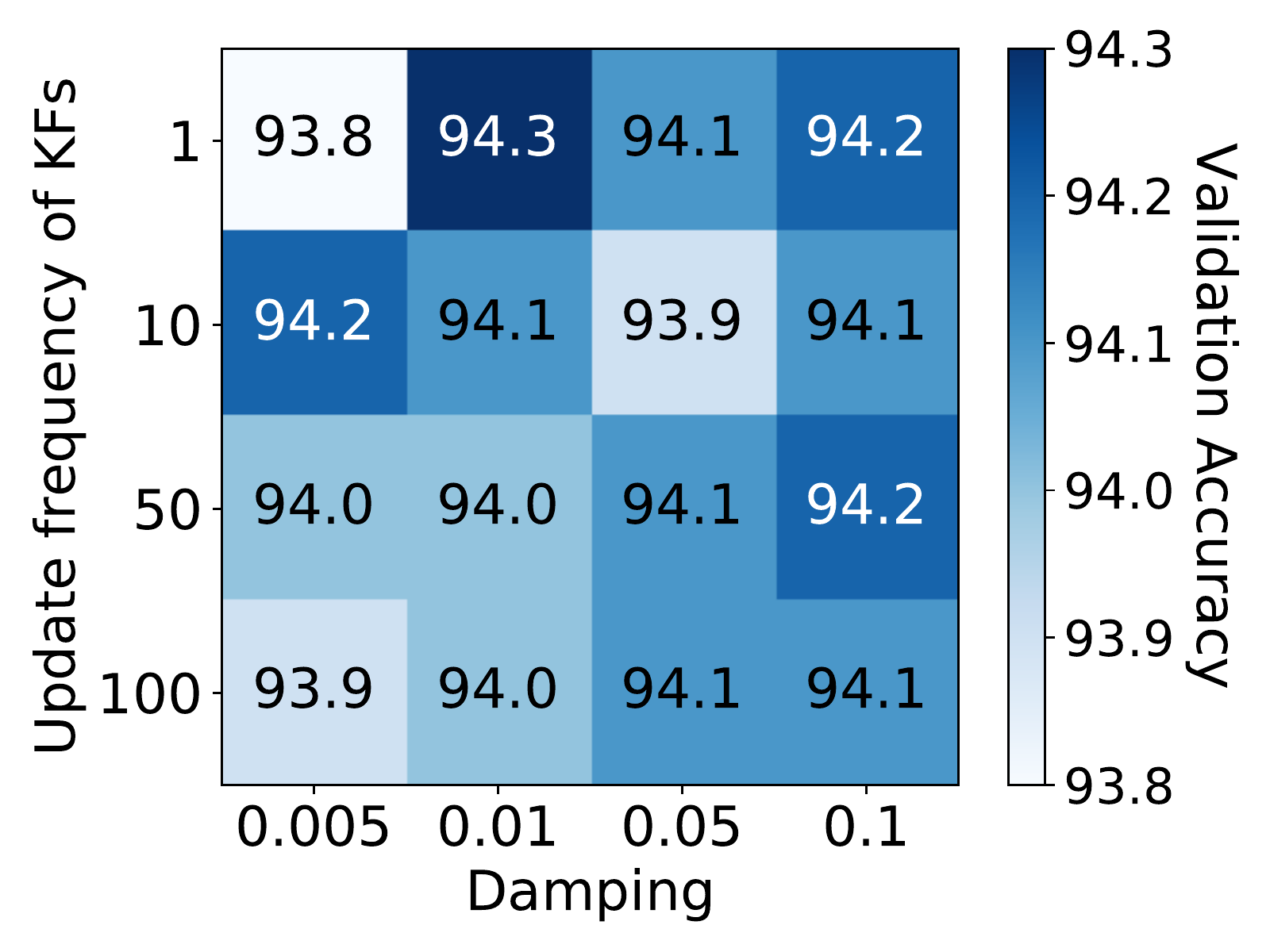}
    }
    \subfloat[DP-KFAC, 4 GPUs]{
        \includegraphics[width=0.48\columnwidth]{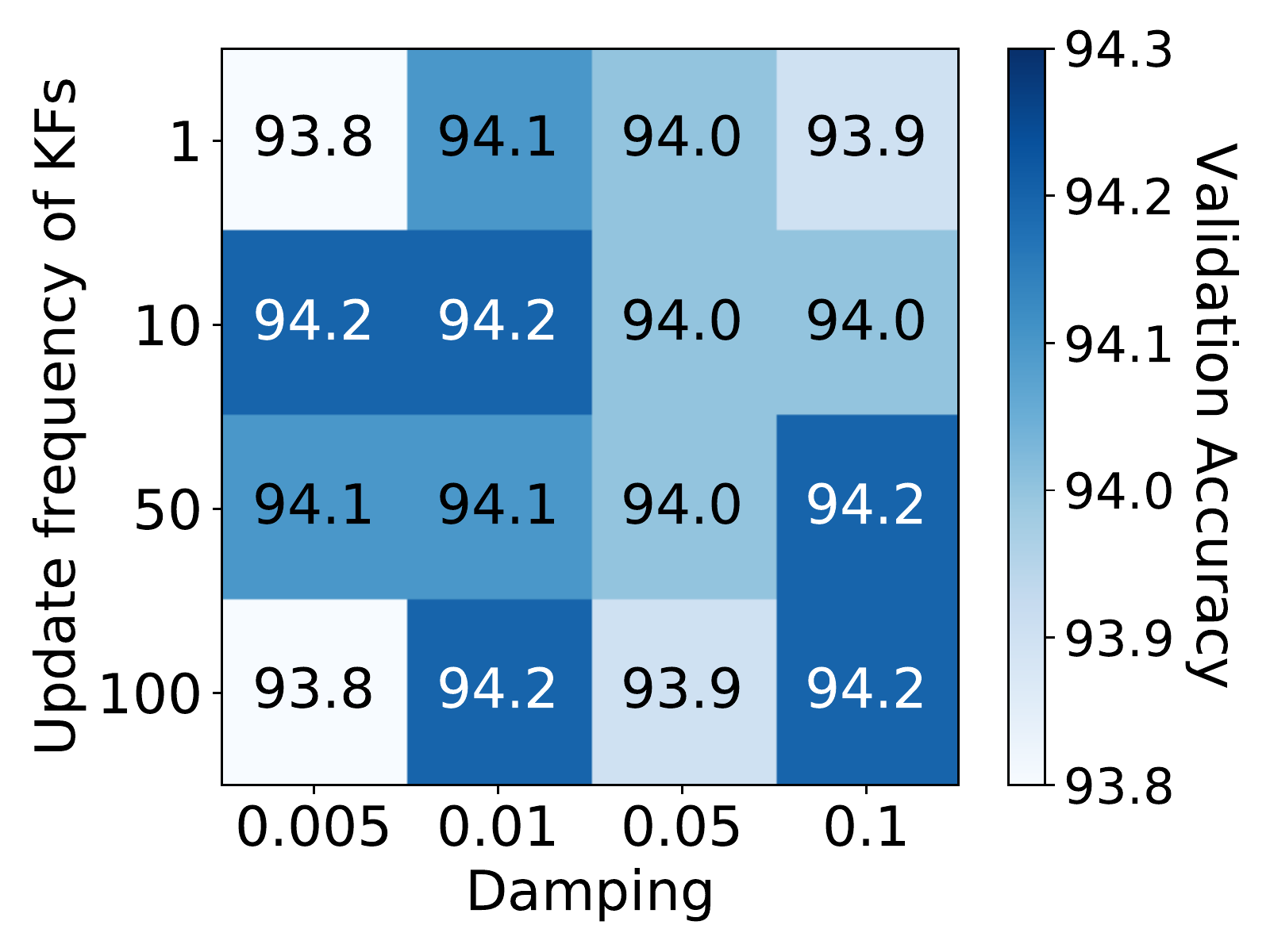}
    } \\
    \subfloat[KAISA, 32 GPUs]{
        \includegraphics[width=0.475\columnwidth]{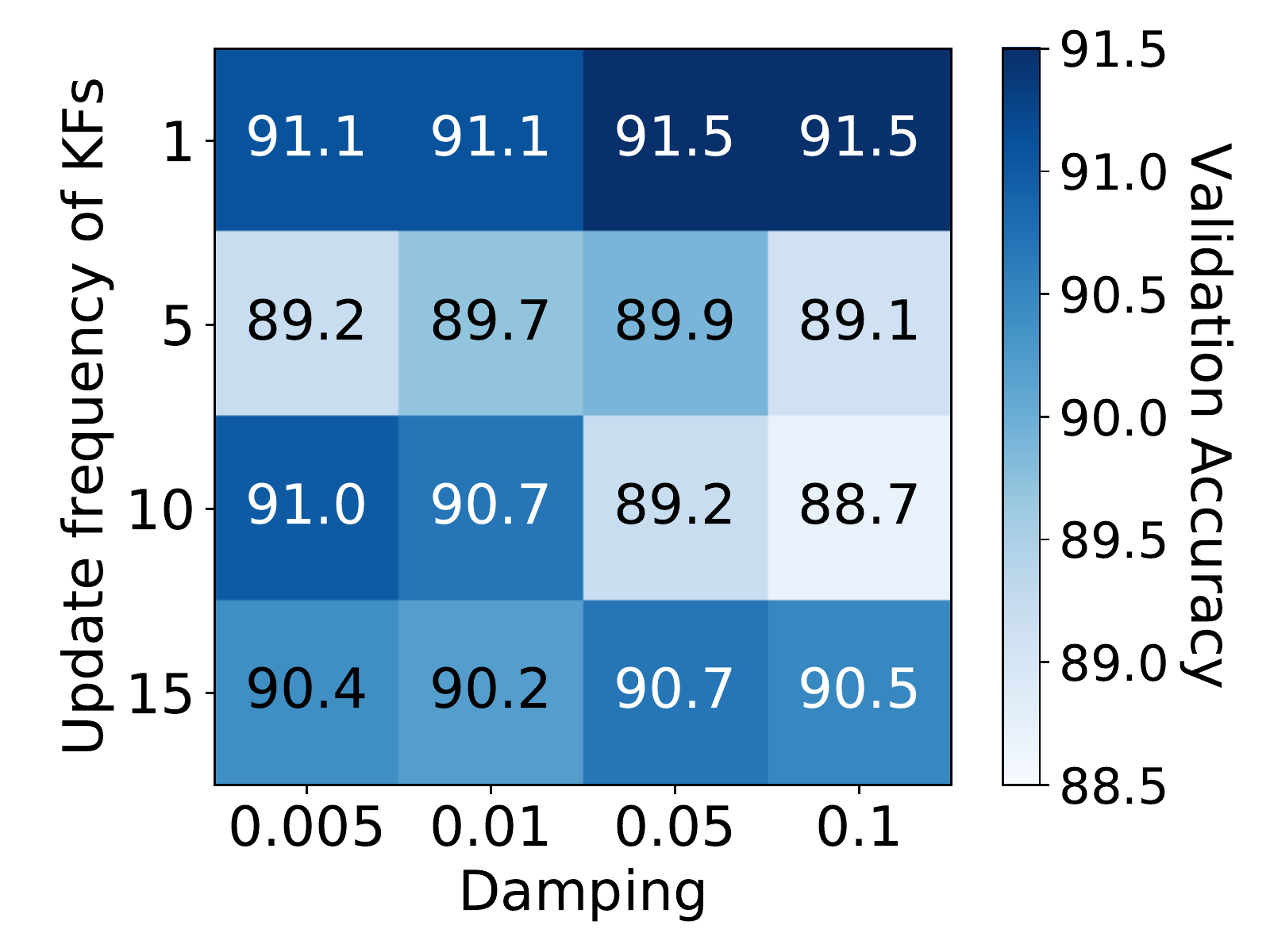}
    }
    \subfloat[DP-KFAC, 32 GPUs]{
        \includegraphics[width=0.475\columnwidth]{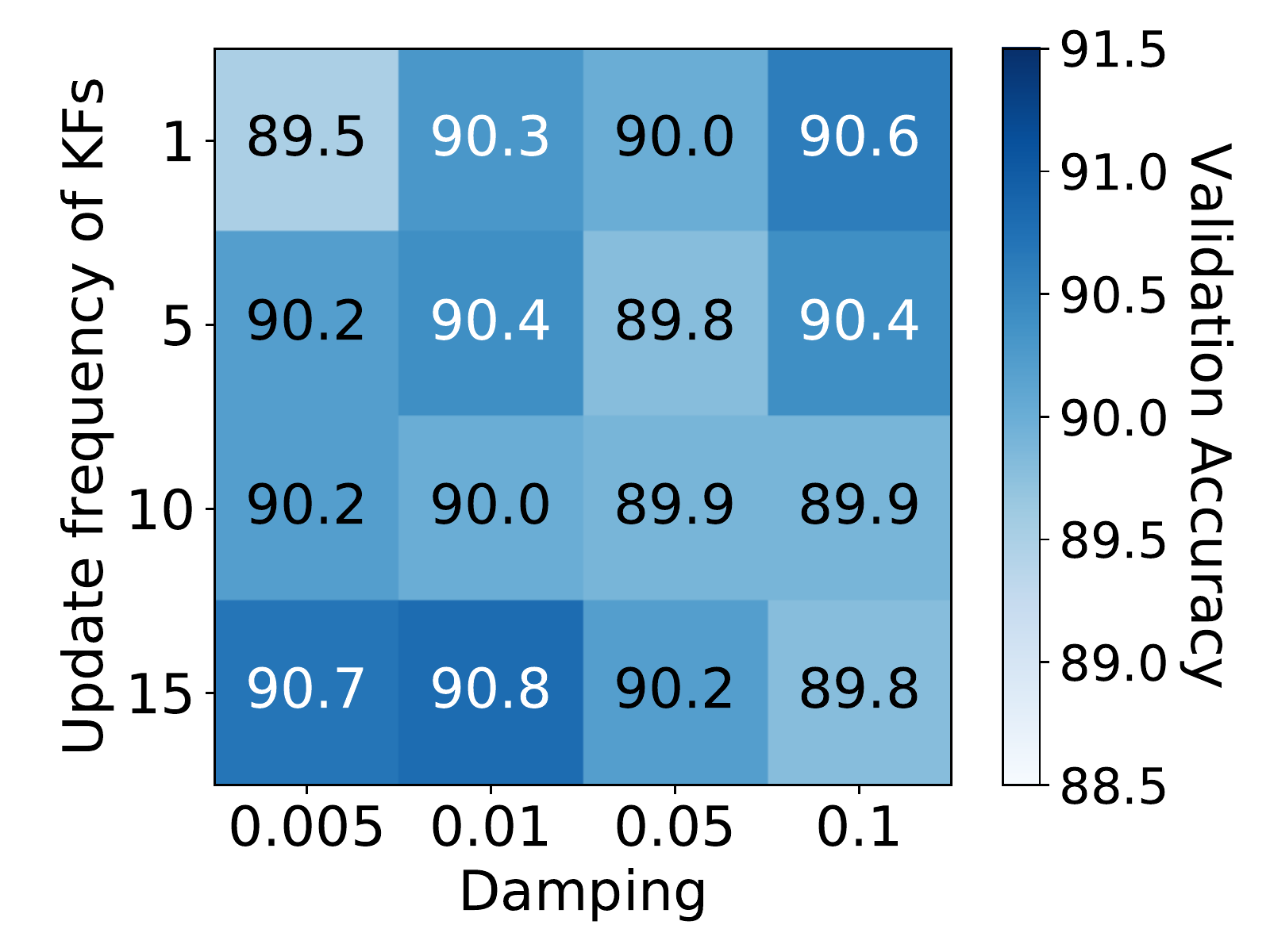}
    }    
    \caption{\change{The effects of damping and the update frequency of KFs on ResNet-110 with KAISA and DP-KFAC.}}
    \label{fig:effect-damping-freq}
\end{figure}
Damping and the update frequency of KFs are two critical hyper-parameters that need to be tuned in K-FAC algorithms~\cite{martens2015optimizing}. Thus, we study their effects on both KAISA and DP-KFAC of training ResNet-110 with \change{4 GPUs and 32 GPUs}. The hyper-parameter tuning spaces and their corresponding validation accuracy are given in Fig.~\ref{fig:effect-damping-freq}. The results show that DP-KFAC can maintain very close performance to KAISA even over different hyper-parameter settings. \change{Specifically, over the 16 runs for both KAISA and DP-KFAC, their average accuracies are 94.07\% and 94.05\% on 4 GPUs, and 90.28\% and 90.17\% on 32 GPUs, respectively. Using 32 GPUs, it shows that the performance drop exists over different hyper-parameters caused by the effects of large-batch training. Furthermore, we find that both KAISA and DP-KFAC can maintain very close accuracy with stale FIM and even large update frequency of KFs (e.g., updating KFs every 100 iterations). This contradicts the observation that skipping FIM updates will affect the convergence performance~\cite{pauloski2020convolutional}, and we contribute it to that the possible performance drop of using stale FIM depends on datasets and models. As it seems that no simple correlation between model accuracy and hyper-parameters (damping and the update frequency of KFs) is exhibited at least for training ResNet-110 on Cifar-10, we leave it as a future work that is worth to be explored.} 


\textbf{Training with Stale FIM.}
\begin{figure}[!t]
    \centering
        \includegraphics[width=0.8\columnwidth]{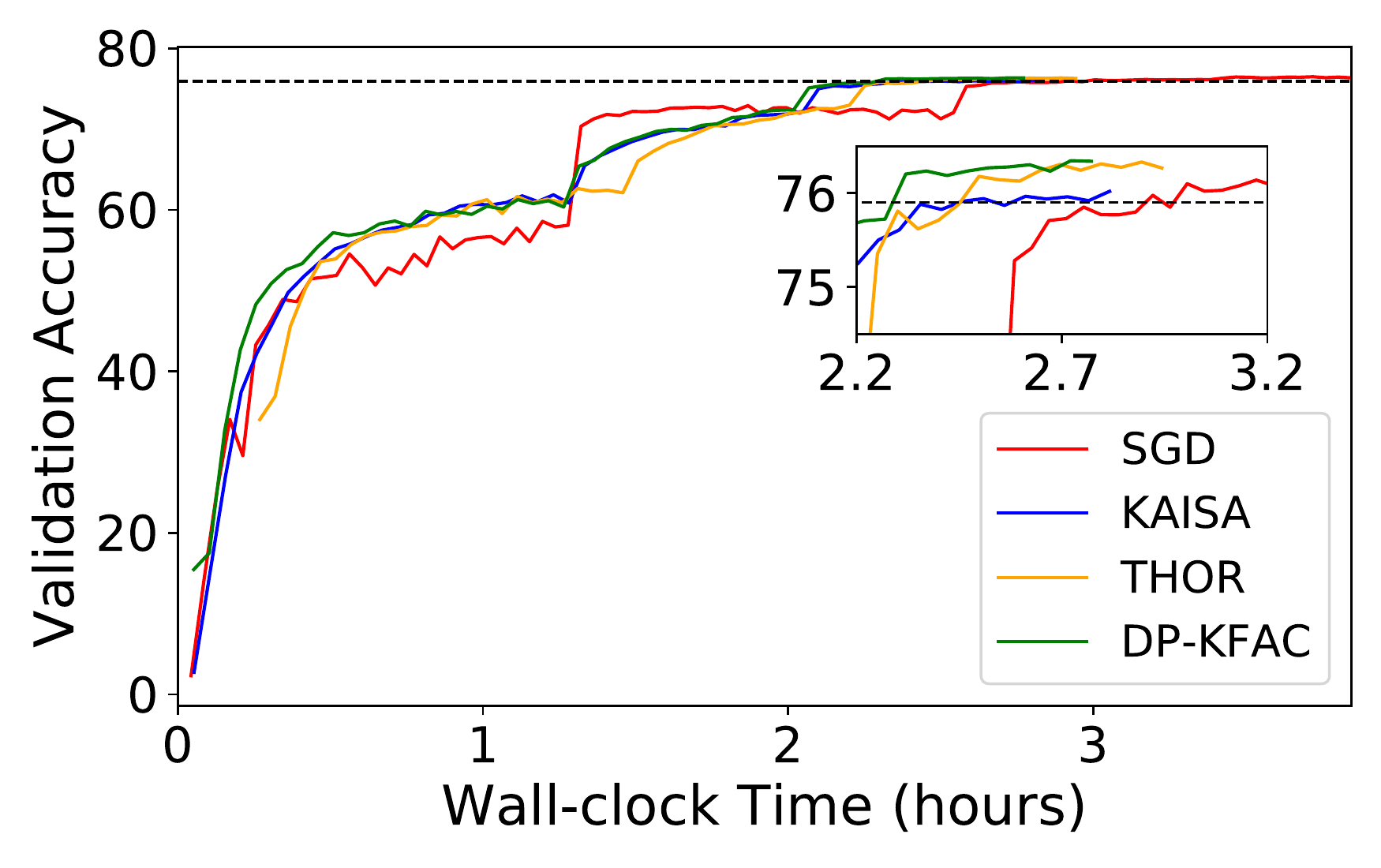}
    \caption{\change{Validation accuracy/loss vs. wall-clock time with stale FIMs on ResNet-50. The black dashed line represents the target accuracy (i.e., 75.9\%). }}
    \label{fig:stale-convergence}
\end{figure}
In practice, it is common to reduce the frequency of FIM updates in K-FAC algorithms~\cite{martens2015optimizing,osawa2019large,pauloski2020convolutional,pauloski2021kaisa,chen2021thor} to accelerate the training process. We hereby verify our DP-KFAC using the stale FIM update with the same configuration as KAISA~\cite{pauloski2021kaisa} using $K\_freq=500$ (the recompute frequency of eigen-decomposition) and $F\_freq=50$ (the update frequency of KFs). We keep the same hyper-parameter settings in all runs as Table~\ref{table:hypers-lr}. We also reproduce THOR~\cite{chen2021thor} for fair comparison on our 64-GPU cluster in training ResNet-50, which utilizes dynamic update intervals ($F\_freq$ and $K\_freq$ are both dynamically increased from 1 to 5000 in the first 10 epochs and $K\_freq=F\_freq=5000$ in the later epochs). The convergence results (top-1 validation accuracy vs. wall-clock time) using stale FIMs are shown in Fig.~\ref{fig:stale-convergence}\footnote{\change{We also conducted experiments with LARS~\cite{you2018imagenet} for better convergence, which requires 70 epochs to reach the target accuracy.}}. It is seen that our DP-KFAC achieves faster convergence performance than both KAISA and THOR in the end-to-end training with stale FIMs. Specifically, DP-KFAC reaches the target accuracy of 75.9\% in just 8350 seconds, which is about 500, 500, and 2450 seconds faster than KAISA, THOR, and SGD, respectively. \change{Therefore, our DP-KFAC is 23\% faster than SGD, while achieving 6\% improvement over KAISA and THOR by reducing the frequency of FIM updates. Though the improvement of DP-KFAC over other D-KFAC algorithms is somehow weakened in this case, we argue that the distributed preconditioning technique are of significance for two reasons: (1) the per-iteration training time optimization is necessary in the cases when skipping KF updates adversely affects the model accuracy, and (2) it can save the GPU memory for storing KF results distributively and the saved memory allows DP-KFAC to train with larger local batch size than KAISA, which can further improve its throughput. }

\change{To verify the benefit of memory saving of DP-KFAC, we compare the system throughput of KAISA (MO) and DP-KFAC (eigen) using different local batch sizes to fully utilize the GPU memory. The results are given in Table~\ref{table:throughput-stale}. It shows that DP-KFAC can use larger local batch size than KAISA with reduced memory footprint, and DP-KFAC achieves an average of 31.3\% speedup over KAISA in the end-to-end training with stale FIM. }

\begin{table}[!t]
    \centering
     \caption{Throughput comparison of different K-FAC algorithms with stale FIM. The results are measured on ImageNet using 64 GPUs. Each algorithm maximizes the local batch size (BS) to fully utilize the GPU memory and updates KFs every 50 iterations (i.e., $K\_freq=F\_freq=50$). }
    \label{table:throughput-stale}
    \centering
    \addtolength{\tabcolsep}{-0.4pt}
    \begin{tabular}{|c|c|c|c|c|c|}
    \hline
   \multirow{2}{*}{Model} & \multicolumn{2}{c|}{KAISA} & \multicolumn{2}{c|}{DP-KFAC} & \multirow{2}{*}{Speedup} \\\cline{2-5} 
   & BS & Throughput & BS & Throughput &  \\\cline{2-6}\hline
   ResNet-50 & 50 & 7729.5 & 60 & 9099.5 & 1.18x \\\hline
   DenseNet-201 & 24 & 2160.3 & 32 & 3245.6 & 1.50x \\\hline
   Inception-v4 & 40 & 3956.7 & 50 & 4984.4 & 1.26x \\\hline
    \end{tabular}
\end{table}

\section{Related Work}\label{sec:related}
Recent years there have been some attempts trying to make second-order optimizations in DNN training be practical~\cite{martens2010deep,martens2015optimizing,bottou2018optimization,yuan2020convergence,thomas2020interplay,yao2021adahessian}, among which the K-FAC based~\cite{martens2015optimizing} algorithms have been successfully applied in large-scale training~\cite{osawa2019large,ueno2020rich,pauloski2021kaisa,chen2021thor}. We mainly introduce the K-FAC related studies. 

\subsection{K-FAC Algorithms}
The second-order algorithms utilize the curvature information to precondition the first-order gradient, which often leads to faster optimizations~\cite{bottou2018optimization}. However, the preconditioning matrix like FIM is too big to construct and invert in DNNs. 
In~\cite{martens2015optimizing}, the K-FAC algorithm is introduced to approximate the FIMs layer-wisely using Kronecker factors, which provides a relatively efficient way to approximate FIMs~\cite{martens2015optimizing,grosse2016kronecker,martens2018kronecker}. In~\cite{george2018fast,gao2021trace}, alternative factored methods have been proposed to further reduce the approximation error of FIMs. Alternatively, in~\cite{tang2021skfac,soori2021tengrad}, the authors propose more efficient approximations with faster inversion. These works do not focus on the distributed optimization, but they are orthogonal to our work. 

\subsection{D-KFAC Algorithms}
The distributed optimization algorithms using K-FAC (D-KFAC)~\cite{osawa2019large,osawa2020scalable,ueno2020rich,pauloski2020convolutional,pauloski2021kaisa} enable the second-order optimization to converge faster than the SGD counterpart on distributed GPU clusters. The implementations of D-KFAC algorithms span from the centralized architecture~\cite{ba2017distributed} to the decentralized architecture~\cite{osawa2019large,osawa2020scalable,ueno2020rich,pauloski2020convolutional,pauloski2021kaisa,chen2021thor}. In the decentralized architecture, it typically uses model parallelism to invert or eigen-decompose~\cite{pauloski2020convolutional,pauloski2021kaisa} KFs of different layers on different workers, and it is able to train the large-scale models faster than SGD in the the end-to-end wall-clock time on GPU clusters. There are also some communication scheduling strategies~\cite{shi2021accelerating,chen2021thor} being proposed to alleviate the communication overheads in D-KFAC through pipelining techniques~\cite{shi2021accelerating} or using dynamic update strategies~\cite{chen2021thor} to improve the scaling efficiency of the distributed system. In this paper, our work is aligned with this direction. 

\section{Conclusion}\label{sec:conclusion}
In this paper, we first analysed the computation and communication bottlenecks in the existing distributed K-FAC (D-KFAC) algorithms. We then introduced our DP-KFAC with distributed preconditioning (DP) that distributes the preconditioning tasks of different layers of a DNN to different GPUs to accelerate training. DP-KFAC enjoys the similar convergence property as the existing D-KFAC algorithms, which is well verified through extensive experiments. With DP-KFAC, the computation, communication, and memory costs can be reduced without sacrificing the convergence performance. We conducted extensive experiments with various modern CNNs and Transformers on different datasets on a 64-GPU cluster. The experimental results showed that our DP-KFAC is much more time-and-memory efficient than state-of-the-art methods without affecting the model accuracy.



\ifCLASSOPTIONcaptionsoff
  \newpage
\fi


\bibliographystyle{IEEEtran}
\bibliography{main}

\begin{thebibliography}{10}
\providecommand{\url}[1]{#1}
\csname url@samestyle\endcsname
\providecommand{\newblock}{\relax}
\providecommand{\bibinfo}[2]{#2}
\providecommand{\BIBentrySTDinterwordspacing}{\spaceskip=0pt\relax}
\providecommand{\BIBentryALTinterwordstretchfactor}{4}
\providecommand{\BIBentryALTinterwordspacing}{\spaceskip=\fontdimen2\font plus
\BIBentryALTinterwordstretchfactor\fontdimen3\font minus
  \fontdimen4\font\relax}
\providecommand{\BIBforeignlanguage}[2]{{%
\expandafter\ifx\csname l@#1\endcsname\relax
\typeout{** WARNING: IEEEtran.bst: No hyphenation pattern has been}%
\typeout{** loaded for the language `#1'. Using the pattern for}%
\typeout{** the default language instead.}%
\else
\language=\csname l@#1\endcsname
\fi
#2}}
\providecommand{\BIBdecl}{\relax}
\BIBdecl

\bibitem{goyal2017accurate}
P.~Goyal, P.~Doll{\'a}r, R.~Girshick, P.~Noordhuis, L.~Wesolowski, A.~Kyrola,
  A.~Tulloch, Y.~Jia, and K.~He, ``Accurate, large minibatch {SGD}: Training
  {ImageNet} in 1 hour,'' \emph{arXiv preprint arXiv:1706.02677}, 2017.

\bibitem{zhang2017poseidon}
H.~Zhang, Z.~Zheng, S.~Xu, W.~Dai, Q.~Ho, X.~Liang, Z.~Hu, J.~Wei, P.~Xie, and
  E.~P. Xing, ``Poseidon: An efficient communication architecture for
  distributed deep learning on {GPU} clusters,'' in \emph{2017 USENIX Annual
  Technical Conference (USENIX ATC 17)}, 2017, pp. 181--193.

\bibitem{jia2018highly}
X.~Jia, S.~Song, S.~Shi, W.~He, Y.~Wang, H.~Rong, F.~Zhou, L.~Xie, Z.~Guo,
  Y.~Yang, L.~Yu, T.~Chen, G.~Hu, and X.~Chu, ``Highly scalable deep learning
  training system with mixed-precision: Training {ImageNet} in four minutes,''
  in \emph{Proc. of Workshop on Systems for ML and Open Source Software,
  collocated with NeurIPS 2018}, 2018.

\bibitem{peng2019generic}
Y.~Peng, Y.~Zhu, Y.~Chen, Y.~Bao, B.~Yi, C.~Lan, C.~Wu, and C.~Guo, ``A generic
  communication scheduler for distributed {DNN} training acceleration,'' in
  \emph{Proceedings of the 27th ACM Symposium on Operating Systems Principles},
  2019, pp. 16--29.

\bibitem{you2020large}
Y.~You, J.~Li, S.~Reddi, J.~Hseu, S.~Kumar, S.~Bhojanapalli, X.~Song,
  J.~Demmel, K.~Keutzer, and C.-J. Hsieh, ``Large batch optimization for deep
  learning: Training {BERT} in 76 minutes,'' in \emph{International Conference
  on Learning Representations}, 2020.

\bibitem{keskar2016large}
N.~S. Keskar, D.~Mudigere, J.~Nocedal, M.~Smelyanskiy, and P.~T.~P. Tang, ``On
  large-batch training for deep learning: Generalization gap and sharp
  minima,'' in \emph{5th International Conference on Learning Representations,
  {ICLR} 2017, Toulon, France, April 24-26, 2017, Conference Track
  Proceedings}, 2017.

\bibitem{hoffer2017train}
E.~Hoffer, I.~Hubara, and D.~Soudry, ``Train longer, generalize better: closing
  the generalization gap in large batch training of neural networks,'' in
  \emph{Advances in Neural Information Processing Systems}, 2017, pp.
  1731--1741.

\bibitem{shallue2019measuring}
C.~J. Shallue, J.~Lee, J.~Antognini, J.~Sohl-Dickstein, R.~Frostig, and G.~E.
  Dahl, ``Measuring the effects of data parallelism on neural network
  training,'' \emph{Journal of Machine Learning Research}, vol.~20, pp. 1--49,
  2019.

\bibitem{amari1998natural}
S.-I. Amari, ``Natural gradient works efficiently in learning,'' \emph{Neural
  computation}, vol.~10, no.~2, pp. 251--276, 1998.

\bibitem{bottou2018optimization}
L.~Bottou, F.~E. Curtis, and J.~Nocedal, ``Optimization methods for large-scale
  machine learning,'' \emph{Siam Review}, vol.~60, no.~2, pp. 223--311, 2018.

\bibitem{zhang2019fast}
G.~Zhang, J.~Martens, and R.~B. Grosse, ``Fast convergence of natural gradient
  descent for over-parameterized neural networks,'' in \emph{NeurIPS}, 2019.

\bibitem{martens2020new}
J.~Martens, ``New insights and perspectives on the natural gradient method,''
  \emph{Journal of Machine Learning Research}, vol.~21, pp. 1--76, 2020.

\bibitem{osawa2019large}
K.~Osawa, Y.~Tsuji, Y.~Ueno, A.~Naruse, R.~Yokota, and S.~Matsuoka,
  ``Large-scale distributed second-order optimization using kronecker-factored
  approximate curvature for deep convolutional neural networks,'' in
  \emph{Proceedings of the IEEE Conference on Computer Vision and Pattern
  Recognition}, 2019, pp. 12\,359--12\,367.

\bibitem{osawa2020scalable}
K.~Osawa, Y.~Tsuji, Y.~Ueno, A.~Naruse, C.~Foo, and R.~Yokota, ``Scalable and
  practical natural gradient for large-scale deep learning,'' \emph{IEEE
  Transactions on Pattern Analysis \& Machine Intelligence}, 2020.

\bibitem{ueno2020rich}
Y.~Ueno, K.~Osawa, Y.~Tsuji, A.~Naruse, and R.~Yokota, ``Rich information is
  affordable: A systematic performance analysis of second-order optimization
  using {K-FAC},'' in \emph{Proceedings of the 26th ACM SIGKDD International
  Conference on Knowledge Discovery \& Data Mining}, 2020, pp. 2145--2153.

\bibitem{pauloski2020convolutional}
J.~G. Pauloski, Z.~Zhang, L.~Huang, W.~Xu, and I.~T. Foster, ``Convolutional
  neural network training with distributed {K-FAC},'' in \emph{Proceedings of
  the International Conference for High Performance Computing, Networking,
  Storage and Analysis}, 2020, pp. 1--14.

\bibitem{pauloski2021kaisa}
J.~G. Pauloski, Q.~Huang, L.~Huang, S.~Venkataraman, K.~Chard, I.~Foster, and
  Z.~Zhang, ``Kaisa: An adaptive second-order optimizer framework for deep
  neural networks,'' in \emph{Proceedings of the International Conference for
  High Performance Computing, Networking, Storage and Analysis}, 2021.

\bibitem{chen2021thor}
M.~Chen, K.~Gao, X.~Liu, Z.~Wang, N.~Ni, Q.~Zhang, L.~Chen, C.~Ding, Z.~Huang,
  M.~Wang \emph{et~al.}, ``{THOR}, trace-based hardware-driven layer-oriented
  natural gradient descent computation,'' in \emph{Proceedings of the AAAI
  Conference on Artificial Intelligence}, vol.~35, no.~8, 2021, pp. 7046--7054.

\bibitem{amari2020does}
S.-i. Amari, J.~Ba, R.~B. Grosse, X.~Li, A.~Nitanda, T.~Suzuki, D.~Wu, and
  J.~Xu, ``When does preconditioning help or hurt generalization?'' in
  \emph{International Conference on Learning Representations}, 2021.

\bibitem{martens2015optimizing}
J.~Martens and R.~Grosse, ``Optimizing neural networks with kronecker-factored
  approximate curvature,'' in \emph{International conference on machine
  learning}, 2015, pp. 2408--2417.

\bibitem{grosse2016kronecker}
R.~Grosse and J.~Martens, ``A kronecker-factored approximate fisher matrix for
  convolution layers,'' in \emph{International Conference on Machine Learning},
  2016, pp. 573--582.

\bibitem{martens2018kronecker}
J.~Martens, J.~Ba, and M.~Johnson, ``Kronecker-factored curvature
  approximations for recurrent neural networks,'' in \emph{International
  Conference on Learning Representations}, 2018.

\bibitem{george2018fast}
T.~George, C.~Laurent, X.~Bouthillier, N.~Ballas, and P.~Vincent, ``Fast
  approximate natural gradient descent in a kronecker factored eigenbasis,''
  \emph{Advances in Neural Information Processing Systems}, vol.~31, pp.
  9550--9560, 2018.

\bibitem{ba2017distributed}
J.~Ba, R.~Grosse, and J.~Martens, ``Distributed second-order optimization using
  kronecker-factored approximations,'' in \emph{International Conference on
  Learning Representations}, 2017.

\bibitem{he2016deep}
K.~He, X.~Zhang, S.~Ren, and J.~Sun, ``Deep residual learning for image
  recognition,'' in \emph{Proceedings of the IEEE conference on computer vision
  and pattern recognition}, 2016, pp. 770--778.

\bibitem{deng2009imagenet}
J.~Deng, W.~Dong, R.~Socher, L.-J. Li, K.~Li, and L.~Fei-Fei, ``{ImageNet}: A
  large-scale hierarchical image database,'' in \emph{2009 IEEE conference on
  computer vision and pattern recognition}.\hskip 1em plus 0.5em minus
  0.4em\relax Ieee, 2009, pp. 248--255.

\bibitem{mlperf}
``Mlperf training v1.1 results,''
  \url{https://mlcommons.org/en/news/mlperf-training-v11/}, accessed:
  2022-03-15.

\bibitem{krizhevsky2009learning}
A.~Krizhevsky, ``Learning multiple layers of features from tiny images,''
  \emph{Citeseer}, 2009.

\bibitem{shi2021accelerating}
S.~Shi, L.~Zhang, and B.~Li, ``Accelerating distributed k-fac with smart
  parallelism of computing and communication tasks,'' in \emph{2021 IEEE 41st
  International Conference on Distributed Computing Systems (ICDCS)}.\hskip 1em
  plus 0.5em minus 0.4em\relax IEEE, 2021.

\bibitem{li2014scaling}
M.~Li, D.~G. Andersen, J.~W. Park, A.~J. Smola, A.~Ahmed, V.~Josifovski,
  J.~Long, E.~J. Shekita, and B.-Y. Su, ``Scaling distributed machine learning
  with the parameter server,'' in \emph{11th USENIX OSDI}, 2014, pp. 583--598.

\bibitem{paszke2019pytorch}
A.~Paszke, S.~Gross, F.~Massa, A.~Lerer, J.~Bradbury, G.~Chanan, T.~Killeen,
  Z.~Lin, N.~Gimelshein, L.~Antiga \emph{et~al.}, ``Pytorch: An imperative
  style, high-performance deep learning library,'' \emph{Advances in neural
  information processing systems}, vol.~32, 2019.

\bibitem{sergeev2018horovod}
A.~Sergeev and M.~Del~Balso, ``Horovod: fast and easy distributed deep learning
  in tensorflow,'' \emph{arXiv preprint arXiv:1802.05799}, 2018.

\bibitem{kingma2014adam}
D.~P. Kingma and J.~Ba, ``Adam: {A} method for stochastic optimization,'' in
  \emph{3rd International Conference on Learning Representations, {ICLR} 2015,
  San Diego, CA, USA, May 7-9, 2015, Conference Track Proceedings}, 2015.

\bibitem{simonyan2014very}
K.~Simonyan and A.~Zisserman, ``Very deep convolutional networks for
  large-scale image recognition,'' \emph{arXiv preprint arXiv:1409.1556}, 2014.

\bibitem{vaswani2017attention}
A.~Vaswani, N.~Shazeer, N.~Parmar, J.~Uszkoreit, L.~Jones, A.~N. Gomez,
  {\L}.~Kaiser, and I.~Polosukhin, ``Attention is all you need,'' in
  \emph{Advances in neural information processing systems}, 2017, pp.
  5998--6008.

\bibitem{elliott2016multi30k}
D.~Elliott, S.~Frank, K.~Sima'an, and L.~Specia, ``Multi30k: Multilingual
  english-german image descriptions,'' in \emph{5th Workshop on Vision and
  Language}.\hskip 1em plus 0.5em minus 0.4em\relax Association for
  Computational Linguistics (ACL), 2016, pp. 70--74.

\bibitem{devlin2019bert}
J.~Devlin, M.-W. Chang, K.~Lee, and K.~Toutanova, ``{BERT}: Pre-training of
  deep bidirectional transformers for language understanding,'' in
  \emph{Proceedings of the 2019 Conference of the North American Chapter of the
  Association for Computational Linguistics: Human Language Technologies,
  Volume 1 (Long and Short Papers)}, 2019, pp. 4171--4186.

\bibitem{Rajpurkar2016SQuAD}
P.~Rajpurkar, J.~Zhang, K.~Lopyrev, and P.~Liang, ``Squad: 100,000+ questions
  for machine comprehension of text,'' in \emph{EMNLP}, 2016.

\bibitem{Wolf2020Transformers}
T.~Wolf, L.~Debut, V.~Sanh, J.~Chaumond, C.~Delangue, A.~Moi, P.~Cistac,
  T.~Rault, R.~Louf, M.~Funtowicz, and J.~Brew, ``Transformers:
  State-of-the-art natural language processing,'' in \emph{EMNLP}, 2020.

\bibitem{huang2017densely}
G.~Huang, Z.~Liu, L.~Van Der~Maaten, and K.~Q. Weinberger, ``Densely connected
  convolutional networks,'' in \emph{Proceedings of the IEEE conference on
  computer vision and pattern recognition}, 2017, pp. 4700--4708.

\bibitem{szegedy2017inception}
C.~Szegedy, S.~Ioffe, V.~Vanhoucke, and A.~A. Alemi, ``Inception-v4,
  inception-resnet and the impact of residual connections on learning,'' in
  \emph{Proc. of The 31st AAAI}, 2017.

\bibitem{lin2020extrapolation}
T.~Lin, L.~Kong, S.~Stich, and M.~Jaggi, ``Extrapolation for large-batch
  training in deep learning,'' in \emph{International Conference on Machine
  Learning}.\hskip 1em plus 0.5em minus 0.4em\relax PMLR, 2020, pp. 6094--6104.

\bibitem{nguyen2022globally}
T.~T. Nguyen, F.~Trahay, J.~Domke, A.~Drozd, E.~Vatai, J.~Liao, M.~Wahib, and
  B.~Gerofi, ``Why globally re-shuffle? revisiting data shuffling in large
  scale deep learning,'' in \emph{IEEE International Parallel \& Distributed
  Processing Symposium}, 2022.

\bibitem{you2018imagenet}
Y.~You, Z.~Zhang, C.-J. Hsieh, J.~Demmel, and K.~Keutzer, ``{ImageNet} training
  in minutes,'' in \emph{Proceedings of the 47th International Conference on
  Parallel Processing}, 2018, pp. 1--10.

\bibitem{martens2010deep}
J.~Martens \emph{et~al.}, ``Deep learning via hessian-free optimization.'' in
  \emph{ICML}, vol.~27, 2010, pp. 735--742.

\bibitem{yuan2020convergence}
X.-T. Yuan and P.~Li, ``On convergence of distributed approximate newton
  methods: Globalization, sharper bounds and beyond.'' \emph{J. Mach. Learn.
  Res.}, vol.~21, pp. 206--1, 2020.

\bibitem{thomas2020interplay}
V.~Thomas, F.~Pedregosa, B.~Merri{\"e}nboer, P.-A. Manzagol, Y.~Bengio, and
  N.~Le~Roux, ``On the interplay between noise and curvature and its effect on
  optimization and generalization,'' in \emph{International Conference on
  Artificial Intelligence and Statistics}.\hskip 1em plus 0.5em minus
  0.4em\relax PMLR, 2020, pp. 3503--3513.

\bibitem{yao2021adahessian}
Z.~Yao, A.~Gholami, S.~Shen, M.~Mustafa, K.~Keutzer, and M.~Mahoney,
  ``Adahessian: An adaptive second order optimizer for machine learning,'' in
  \emph{Proceedings of the AAAI Conference on Artificial Intelligence},
  vol.~35, no.~12, 2021, pp. 10\,665--10\,673.

\bibitem{gao2021trace}
K.~Gao, X.~Liu, Z.~Huang, M.~Wang, Z.~Wang, D.~Xu, and F.~Yu, ``A
  trace-restricted kronecker-factored approximation to natural gradient,'' in
  \emph{Proceedings of the AAAI Conference on Artificial Intelligence},
  vol.~35, no.~9, 2021, pp. 7519--7527.

\bibitem{tang2021skfac}
Z.~Tang, F.~Jiang, M.~Gong, H.~Li, Y.~Wu, F.~Yu, Z.~Wang, and M.~Wang, ``Skfac:
  Training neural networks with faster kronecker-factored approximate
  curvature,'' in \emph{Proceedings of the IEEE/CVF Conference on Computer
  Vision and Pattern Recognition}, 2021, pp. 13\,479--13\,487.

\bibitem{soori2021tengrad}
S.~Soori, B.~Can, B.~Mu, M.~G{\"u}rb{\"u}zbalaban, and M.~M. Dehnavi,
  ``Tengrad: Time-efficient natural gradient descent with exact fisher-block
  inversion,'' \emph{arXiv preprint arXiv:2106.03947}, 2021.

\end{thebibliography}

\begin{IEEEbiography}[{\includegraphics[width=1in,height=1.25in,clip,keepaspectratio]{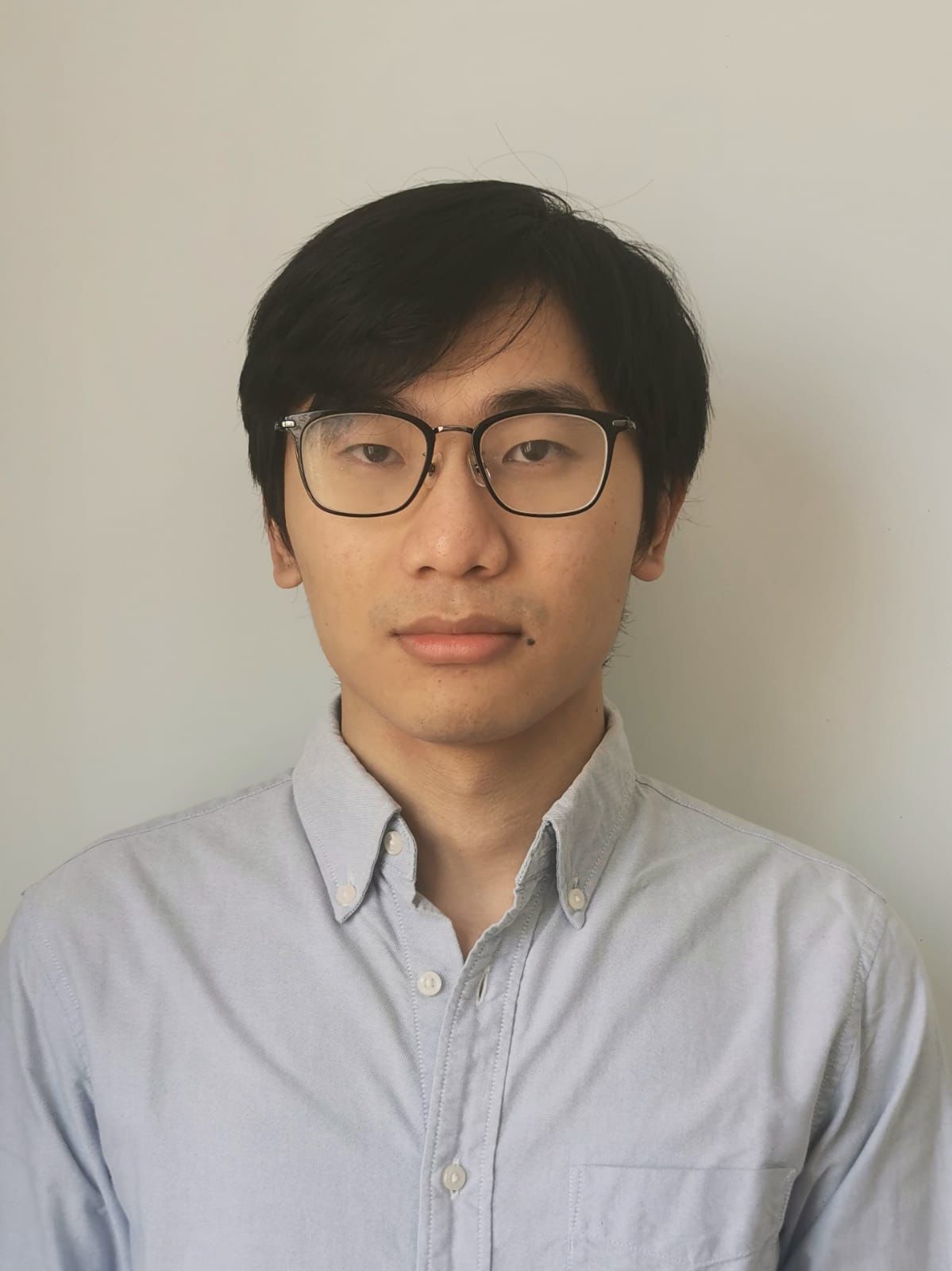}}]{Lin Zhang} received the B.S. degree at School of Electrical Engineering and Automation from Zhejiang University in 2018. He is currently pursuing the Ph.D. degree in the Department of Computer Science and Engineering at the Hong Kong University of Science and Technology. His research interests include machine learning systems and applications, with a special focus on second-order optimization methods, and unsupervised learning on graphs. 
\end{IEEEbiography}

\begin{IEEEbiography}[{\includegraphics[width=1in,height=1.25in,clip,keepaspectratio]{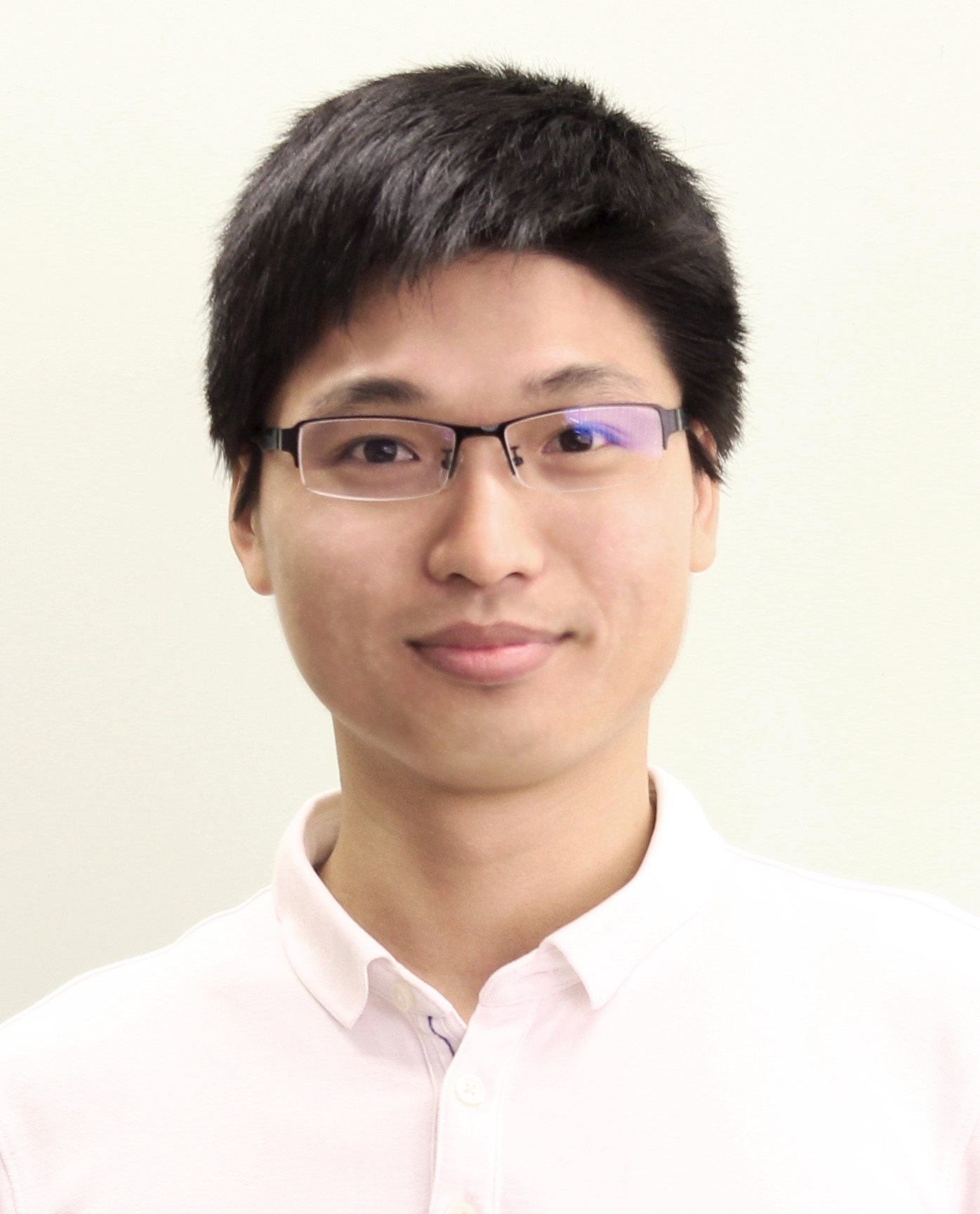}}]{Shaohuai Shi} received a B.E. degree in software engineering from South China University of Technology, P.R. China, in 2010, an MS degree in computer science from Harbin Institute of Technology, P.R. China in 2013, and a Ph.D. degree in computer science from Hong Kong Baptist University in 2020. He is currently a research assistant professor in the Department of Computer Science and Engineering at the Hong
Kong University of Science and Technology. His research interests include GPU computing and machine learning systems. He is a member of the IEEE.
\end{IEEEbiography}

\begin{IEEEbiography}[{\includegraphics[width=1in,height=1.25in,clip,keepaspectratio]{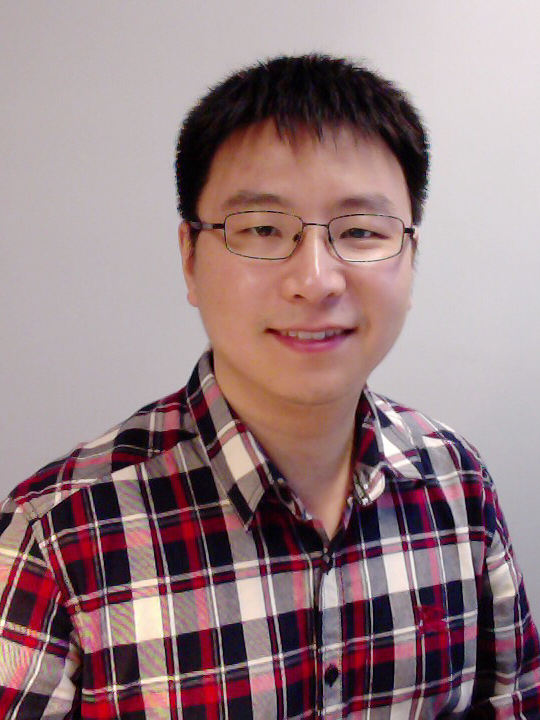}}]{Wei Wang} received his B.Engr. and M.Engr. degrees from the Department of Electrical Engineering, Shanghai Jiao Tong University, China, in 2007 and 2010, respectively and his Ph.D. degree from the Department of Electrical and Computer Engineering, University of Toronto, Canada, in 2015. Since 2015, he has been with the Department of Computer Science and Engineering at the Hong Kong University of Science and Technology (HKUST), where he is currently an Associate Professor. He is also affiliated with the HKUST Big Data Institute. Dr. Wang’s research interests cover the broad area of distributed systems, with focus on serverless computing, machine learning systems, and cloud resource management. He published extensively in the premier conferences and journals of his fields. His research has won the Best Paper Runner Up awards of IEEE ICDCS 2021 and USENIX ICAC 2013.
\end{IEEEbiography}

\begin{IEEEbiography}[{\includegraphics[width=1in,height=1.25in,clip,keepaspectratio]{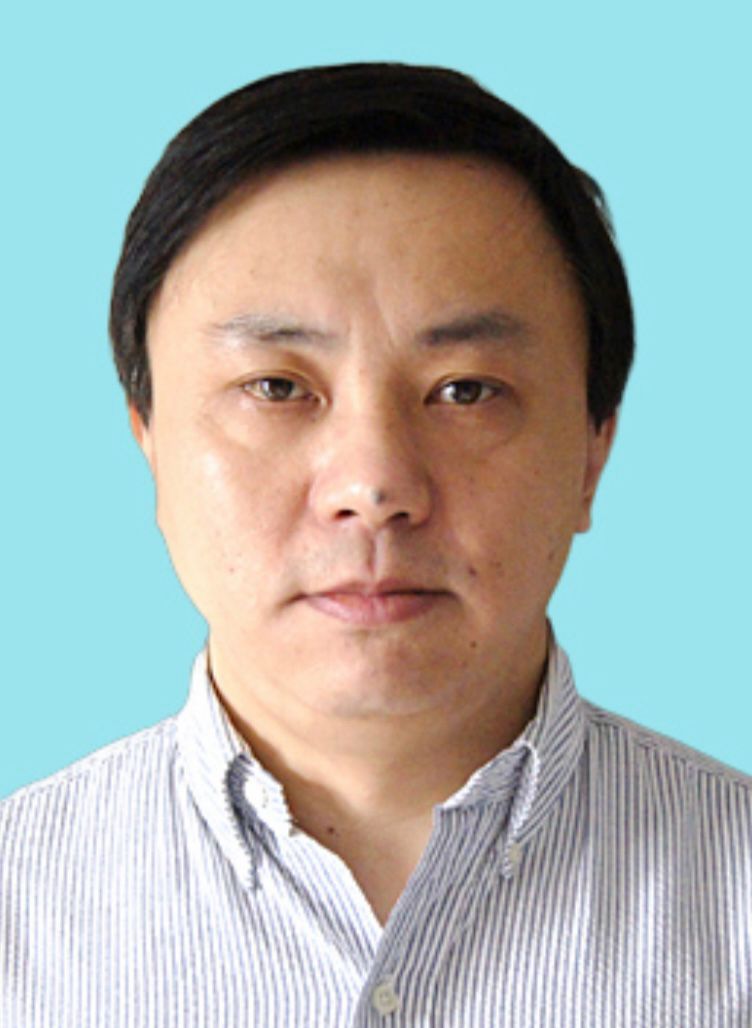}}]{Bo Li} is a professor in the Department of Computer Science and Engineering, Hong Kong University of Science and Technology. He holds the Cheung Kong chair professor in Shanghai Jiao Tong University. Prior to that, he was with IBM Networking System Division, Research Triangle Park, North Carolina. He was an adjunct researcher with Microsoft Research Asia-MSRA and was a visiting scientist in Microsoft Advanced Technology Center (ATC). He has been a technical advisor for China Cache Corp. (NASDAQ CCIH) since 2007. He is an adjunct professor with the Huazhong University of Science and Technology, Wuhan, China. His recent research interests include: large-scale content distribution in the Internet, Peer-to-Peer media streaming, the Internet topology, cloud computing, green computing and communications. He is a fellow of the IEEE for “contribution to content distributions via the Internet”. He received the Young Investigator Award from the National Natural Science Foundation of China (NSFC) in 2004. He served as a Distinguished lecturer of the IEEE Communications Society (2006-2007). He was a corecipient for three Best Paper Awards from IEEE, and the Best System Track Paper in ACM Multimedia (2009).
\end{IEEEbiography}

\end{document}